%% file: main.tex
\documentclass[acmsmall]{acmart}

\usepackage[utf8]{inputenc}
\usepackage[capitalise]{cleveref}
\crefname{section}{Sec.}{Secs.}
\Crefname{section}{Section}{Sections}
\crefname{table}{Tab.}{Tabs.}
\Crefname{table}{Table}{Tables}
\crefname{figure}{Fig.}{Figs.}
\Crefname{figure}{Figure}{Figures}
\crefname{equation}{Eq.}{Eqs.}
\Crefname{equation}{Equation}{Equations}
\usepackage{booktabs} 
\usepackage{multirow}
\usepackage{float}
\usepackage{makecell}
\usepackage{tabularray}

\usepackage{tikz}
\usepackage{longtable}
\usepackage{tikz-qtree}
\usepackage{geometry}
\usepackage[T1]{fontenc}
\usepackage[edges]{forest}
\usetikzlibrary{trees} %
\usepackage[figuresright]{rotating}

\usepackage{xcolor,colortbl}
\usepackage{pifont}
\usepackage[flushleft]{threeparttable}

\newcommand*{\belowrulesepcolor}[1]{%
  \noalign{%
    \kern-\belowrulesep 
    \begingroup 
      \color{#1}%
      \hrule height\belowrulesep 
    \endgroup 
    \vspace{-0.03mm}
  }%
} 
\newcommand*{\aboverulesepcolor}[1]{%
  \noalign{%
  \vspace{-0.03mm}
    \begingroup 
      \color{#1}%
      \hrule height\aboverulesep 
    \endgroup 
    \kern-\aboverulesep 
  }%
}

\newcommand{\eg}{\emph{e.g.}}

\newcommand{\ie}{\emph{i.e.}}
\newcommand{\etal}{\emph{et al.}}

\newcommand{\z}{\mathbf{z}}
\newcommand{\x}{\mathbf{x}} 
\newcommand{\y}{\mathbf{y}}
\newcommand{\w}{\mathbf{w}}
\newcommand{\n}{\mathbf{n}} 

\newcommand{\bb}{\mathbf{b}}

\newcommand{\A}{\mathbf{A}}

\newcommand{\ncmark}{\ding{51}}
\newcommand{\nxmark}{\ding{55}}

\newcommand{\vspacefigtext}{\vspace{-3mm}}

\newcommand{\revise}{\textcolor{black}} 
\newcommand{\comments}{\textcolor{black}} 

\graphicspath{{image/}}

\input{command/figure.tex}

\newif\ifcomment

\commenttrue

\AtBeginDocument{%
  }
    
\setcopyright{rightsretained}
\acmJournal{CSUR}
\acmYear{2023} \acmVolume{1} \acmNumber{1} \acmArticle{1} \acmMonth{1} \acmPrice{}\acmDOI{10.1145/3626193}

\acmPrice{15.00}
\acmISBN{978-1-4503-XXXX-X/18/06}

\begin{document}

\title{A Survey on Deep Generative 3D-aware Image Synthesis}

\author{Weihao Xia}
\email{weihao.xia.21@ucl.ac.uk}
\orcid{0000-0003-0087-3525}

\author{Jing-Hao Xue}
\email{jinghao.xue@ucl.ac.uk}
\orcid{/0000-0003-1174-610X}
\affiliation{%
\department{Department of Statistical Science}
\institution{University College London}
\city{London}
\country{UK}
}

\renewcommand{\shortauthors}{Xia et al.}

\begin{abstract}
\revise{Recent years have seen remarkable progress in deep learning powered visual content creation. 
This includes deep generative 3D-aware image synthesis, which produces high-fidelity images in a 3D-consistent manner while simultaneously capturing compact surfaces of objects from pure image collections without the need for any 3D supervision, thus bridging the gap between 2D imagery and 3D reality.
The field of computer vision has been recently captivated by the task of deep generative 3D-aware image synthesis, with hundreds of papers appearing in top-tier journals and conferences  over the past few years (mainly the past two years), but there lacks a comprehensive survey of this remarkable and swift progress.
Our survey aims to introduce new researchers to this topic, provide a useful reference for related works, and stimulate future research directions through our discussion section.
Apart from the presented papers, we aim to constantly update the latest relevant papers along with corresponding implementations at
\url{https://weihaox.github.io/3D-aware-Gen}.}
\end{abstract}



\maketitle

{\section{Introduction}\label{sec:intro}}

A tremendous amount of progress has been made in \revise{deep neural networks} that lead to photorealistic image synthesis. Despite achieving compelling results, most approaches focus on two-dimensional (2D) images, overlooking the three-dimensional (3D) nature of the physical world. The lack of 3D structure, therefore, inevitably limits some of their practical applications. 
\revise{Recent studies have thus  proposed generative models that are 3D-aware. That is, they incorporate 3D information into the generative models to enhance control (especially in terms of multiconsistency) over the generated images.
Examples depicted in~\cref{fig:taskillustration} elucidate that the objective is to produce high-quality renderings which maintain consistency across various views.
}
In contrast to the 2D generative models, the recently developed 3D-aware generative models~\cite{gu2022stylenerf,chan2021pigan} bridge the gap between 2D images and 3D physical world. 
The physical world surrounding us is intrinsically three-dimensional and images depict reality under certain conditions of geometry, material, and illumination, making it natural to model the image generation process in 3D spaces.
As shown in~\cref{fig:overview}, classical rendering (a) renders images at certain camera positions given human-designed or scanned 3D shape models; inverse rendering (b) recovers the underlying intrinsic properties of the 3D physical world from 2D images; 2D image generation (c) is mostly driven by generative models, which have achieved impressive results in photorealistic image synthesis; and 3D-aware generative models (d) offers the possibility of replacing the classical rendering pipeline with effective and efficient models learned directly from images. 

\figtaskillustration

Despite striking progress has been made recently in research of deep generative 3D-aware image synthesis, it lacks a timely and systematic review of this progress. In this work, we fill the gap by presenting a comprehensive survey of the latest research in deep generative 3D-aware image synthesis methods. 
We envision that our work will elucidate design considerations and advanced methods for deep generative 3D-aware image synthesis, present its advantages and disadvantages of different kinds, and suggest future research directions. 
\revise{\cref{fig:phylogenetic_tree} provides a structured outline and taxonomy of this survey. \cref{fig:timeline} is a chronological overview of representative deep generative 3D-aware image synthesis methods.
We propose to categorize the deep generative 3D-aware image synthesis methods into two primary categories: 3D control of 2D generative models (\cref{sec:3d_control_of_2d_generative_models}) and 3D-aware generative models from single image collections (\cref{sec:3d_generative_models}).
Then, every category is further divided into some subcategories depending on the experimental setting or the specific utilization of 3D information.
\revise{In particular, 3D control of 2D generative models are further divided into 1) exploring 3D control in 2D latent spaces (\cref{subsec:exploring_3d_control_in_2d_latent_space}), 2) 3D parameters as controls (\cref{subsec:3d_parameter_as_controls}), and 3) 3D priors as constraints (\cref{subsec:3d_prior_as_constraints}).}
\cref{sec:3d_generative_models} summarizes methods that target generating photorealistic and multi-view-consistent images by learning 3D representations from single-view image collections.
Broadly speaking, this category leverages neural 3D representations to represent scenes, use differentiable neural renderers to render them into the image plane, and optimize the network parameters by minimizing the difference between rendered images and observed images.
}

Here, we present a timely up-to-date overview of the growing field of deep generative 3D-aware image synthesis. 
Considering the lack of a comprehensive survey and an increasing interest and popularity, we believe it necessary to organize one to help computer vision practitioners with this emerging topic. 
The purpose of this survey is to provide researchers new to the field with a comprehensive understanding of  deep generative 3D-aware image synthesis methods and show the superior performance over the status quo approaches. To conclude, we highlight several open research directions and problems that need further investigation.
The scope of this fast-expanding field is rather extensive and a panoramic review would be challenging. We shall present only representative methods of deep generative  3D-aware image synthesis rather than listing exhaustively all literature.
This review can therefore serve as a pedagogical tool, providing researchers with the key information about typical methods of  deep generative 3D-aware image synthesis. Researchers can use these general guidelines to develop the most appropriate technique for their own particular study.
The main technical contributions of this work are as follows:
\begin{itemize}
    \item Systematic taxonomy. 
    \revise{We propose a hierarchical taxonomy for  deep generative 3D-aware image synthesis research. We categorize existing models into two main categories: 3D control of 2D generative models and 3D-aware generative models from image collections.}
    \item Comprehensive review. We provide a comprehensive overview of the existing state-of-the-art  deep generative 3D-aware image synthesis methods. 
    We compare and analyze the main characteristics and improvements for each type, assessing their strengths and weaknesses.
    \item Outstanding challenges. %
    We present open research problems and provide some suggestions for the future development of  deep generative 3D-aware image synthesis.
    \item In an attempt to continuously track recent developments in this fast advancing field, we provide an accompanying webpage which catalogs papers addressing  deep generative 3D-aware image synthesis, according to our problem-based taxonomy: \url{https://weihaox.github.io/3D-aware-Gen}.
\end{itemize}

\figoverview

\section{Background}
\label{sec:background}

This section introduces a few important concepts as the background.
In order to formulate deep generative 3D-aware image synthesis, we first clarify how 2D and 3D data are expressed, and how they are converted between each other. 
Moreover, we introduce two key elements involved in most deep generative 3D-aware image synthesis methods: implicit neural representations and differentiable neural rendering.

\subsection{2D and 3D Data, Rendering and Inverse Rendering}
\label{subsec:3d_shape_representation}

The 2D images depict a glimpse into the surrounding 3D physical world with its geometry, materials, and illumination conditions at that moment.
Images are composed of an array of pixels (picture elements).
The 3D reality can be represented in many different ways, each with its own advantages and disadvantages. 
There are several examples of such \textbf{3D shape representations}, including depth images, point clouds, voxel grids, and meshes.
\textbf{Depth images} contain distance information between the object and the camera at every pixel. The distance encodes 3D geometry information from a fixed point of view. Layered depth images (LDIs) use several layers of depth maps and their associated color values to depict a scene.
\textbf{Point clouds} comprise vertices in 3D space, represented by coordinates along the x, y, and z axes. These types of data can be acquired by 3D scanners, such as LiDARs or RGB-D sensors, from one or more viewpoints.
\textbf{Voxel grids} describe a scene or object using a regular grid in 3D space. A voxel (volume element) in 3D space is analogous to a pixel in a 2D image. A voxel grid can be created from a point cloud by voxelization, which groups all 3D points within a voxel.
\textbf{Meshes} are a collection of vertices, edges, and polygonal faces. In contrast to a point cloud, which only provides vertices locations, a mesh also provides surface information of an object. 
Nevertheless, deep learning does not provide a straightforward way to process surface information. Instead, many techniques resort to sampling points from the surfaces to create a point cloud from the mesh representation.

As shown in~\cref{fig:overview}(a), images can be obtained by rendering a 3D object or scene under certain viewpoints and lighting conditions. This forward process is called \textbf{rendering} (image synthesis).
Rendering has been studied in computer graphics and a wide variety of renderers are available for use.
The reverse process, \textbf{inverse rendering}, as shown in~\cref{fig:overview}(b), is to infer underlying intrinsic components of a scene from rendered 2D images. %
These properties include shape (surface, depth, normal), material (albedo, reflectivity, shininess), and lighting (direction, intensity), which can be further used to render photorealistic images.
The inverse rendering papers
are not classified as 3D-aware image synthesis methods in this survey as they are not deliberately designed for this purpose. %
3D-aware image synthesis in this survey include a similar inverse rendering process and a rendering process. 
In contrast, these methods typically do not produce explicit 3D representations such as meshes or voxels for rendering.
They learn~\textbf{neural 3D representations} (mostly implicit functions), render them into images with~\textbf{differentiable neural rendering} technique, and optimize the network parameters by minimizing the difference between the observed and rendered images. 

\input{command/tree.tex}

\subsection{Implicit Scene Representations}

In computer vision and computer graphics, 3D shapes are traditionally represented as explicit representations like depths, voxels, point clouds, or meshes. Recent methods propose to represent 3D scenes with neural implicit functions, such as occupancy field~\cite{mescheder2019occupancy}, signed distance field~\cite{park2019deepsdf}, and radiance field~\cite{mildenhall2020nerf}. The implicit neural representation (INR, neural fields, or coordinate-based representation) is a novel way to parameterize signals across a wide range of domains. Taking images as an example, INR parameterizes an image as a continuous function that maps pixel coordinates to RGB colors. The implicit functions are often not analytically tractable and are hence approximated by neural networks. Here are some popular examples of INR.

\vspace{1mm} 
\textbf{\revise{Neural} Occupancy Field}~\cite{mescheder2019occupancy,peng2020convolutional,niemeyer2020differentiable} implicitly represents a 3D surface with the continuous decision boundary of a neural classifier. 
This function approximated with a neural network assigns to every location $p \in \mathbb{R}^3$ an occupancy probability between 0 and 1.
Given an observation (\eg, image or point cloud) $x \in \mathcal{X}$ and a location $p \in \mathbb{R}^3$, the representation can be simply parameterized by a neural network $f_\theta$ that takes a pair $(p, x)$ as input and outputs a real number which represents the probability of occupancy: $f_\theta: \mathbb{R}^3 \times \mathcal{X} \rightarrow[0,1]$.

\vspace{1mm} 
\textbf{\revise{Neural} Signed Distance Field}~\cite{park2019deepsdf} is a continuous function that models the distance from a queried location to the nearest point on a shape’s surface, whose sign indicates if this location is inside (negative) or outside (positive):
$S D F(\boldsymbol{x})=s, \boldsymbol{x} \in \mathbb{R}^3, s \in \mathbb{R}$.
The underlying surface is implicitly described as the zero iso-surface decision boundaries of feed-forward networks $S D F(\cdot)=0$. This implicit surface can be rendered by raycasting or rasterizing a mesh obtained through marching cubes~\cite{lorensen1987marching}.

\vspace{1mm} 
\textbf{\revise{Neural} Radiance Field}~\cite{mildenhall2020nerf} (NeRF) has attracted growing attention due to its compelling results in novel view synthesis on complex scenes.
It leverages an MLP network to approximate the radiance fields of static 3D scenes and uses the classic volume rendering technique~\cite{kajiya1984ray} to estimate the color of each pixel.
This function takes as input a 3D location $\boldsymbol{x}$ and 2D viewing direction $\boldsymbol{d}$, and outputs an directional emitted RGB color $\boldsymbol{c}$ and volume density $\sigma$: $f_{\theta}: (\boldsymbol{x}, \boldsymbol{d}) \rightarrow(\boldsymbol{c}, \sigma)$.
It captures 3D geometric details based on pure 2D supervision by learning the reconstruction of given views. 

There also exist many other implicit functions proposed to represent a scene, such as neural sparse voxel fields~\cite{liu2020neural}, or neural volumes~\cite{lombardi2019neural}.

\subsection{Differentiable Neural Rendering}

3D rendering is a function that outputs a 2D image from a 3D scene. Differentiable rendering provides a differentiable rendering function, that is, it computes the derivatives of that function in response to different parameters of the scene. Once a renderer is differentiable, it can be integrated into the optimization of neural networks. One use case for differentiable rendering is to compute a loss in rendered image space and back propagation can be applied to train the network. 
Driven by the prevalence of NeRF-based methods~\cite{mildenhall2020nerf}, volume rendering~\cite{kajiya1984ray} becomes the most commonly used differentiable renderer among the methods that this survey targets.
It is naturally differentiable, and the only input required to optimize the NeRF representation is a set of images with known camera poses.
Given volume density and color functions, volume rendering is used to obtain the color $C(\mathbf{r})$ of any camera ray $\mathbf{r}(t)=\mathbf{o}+t \mathbf{d}$, with camera position $\mathbf{o}$ and viewing direction $\mathbf{d}$ using
\begin{equation}
    C(\mathbf{r})=\int_{t_1}^{t_2} T(t) \cdot \sigma(\mathbf{r}(t)) \cdot \mathbf{c}(\mathbf{r}(t), \mathbf{d}) \cdot d t,
    \text { where } T(t)=\exp \left(-\int_{t_1}^t \sigma(\mathbf{r}(s)) d s\right).
\end{equation}
$T(t)$ denotes the accumulated transmittance, representing the probability that the ray travels from $t_1$ to $t$ without being intercepted.
The rendered image can be obtained by tracing the camera rays $C(\mathbf{r})$ through each pixel of the to-be-synthesized image.

\input{command/timeline}

\subsection{\revise{INR-based Novel View Synthesis}}
\label{subsec:3d_novel_view_synthesis}

\revise{Novel view synthesis~\cite{mildenhall2020nerf,yu2021pixelnerf,rematas2022urbannerf,gafni2021dynamic} is a long-standing problem that involves rendering frames of scenes from new camera viewpoints.  
There are existing methods that depend on implicit 3D scene representations. One of the most representative studies in the field is NeRF, which employs neural networks to capture the continuous 3D scene structure within the network weights, resulting in photorealistic synthesis outcomes.
These methods usually operate under the Single-Scene Overfitting (SSO) experiment. 
This approach aims to render novel views by learning a deep neural scene representation from multi-view image collections of a specific scene or object.
These models are trained per-scene and are primarily designed for tasks of 3D reconstruction, novel view synthesis, or free viewpoint rendering. 
Its success has inspired various extensions and improvements, leading to a rich body of work in the NeRF family of methods. These variants aim to address limitations, enhance efficiency, and improve the quality and diversity of the generated images.
For more details on novel view synthesis, please refer to recent methods~\cite{mildenhall2020nerf,yu2021pixelnerf,rematas2022urbannerf,gafni2021dynamic}, surveys~\cite{tewari2022advances}, and the appendix of this survey.}

\revise{The focus of this survey, which is on 3D-aware generative models, bears close relevance to 3D novel view synthesis, particularly the INR-based methods. Many 3D-aware generative models derive inspiration from the field of INR-based 3D novel view synthesis. In the following sections, we will demonstrate how these two areas can benefit each other in solving key issues in their respective research fields, in terms of technical implementation and application scenarios.
}

\subsection{Dataset}
\label{subsec:dataset}

Similar to 2D models, 3D-aware generative models
aim to produce photorealistic and multi-view consistent images. Therefore, the same single-view image datasets are used as in the 2D methods.
\revise{These datasets predominantly consist of single-view image collections, which are usually unstructured and unannotated. These are typically employed by unconditional generative models and some conditional methods.
These image collections can be further categorized into \textit{single} objects and \textit{multiple} objects according to the prominent object numbers in the foreground, and \textit{simple} shape and \textit{variable} shape according to the fineness of the object.
Most datasets pertain to a specific category. Current popular categories include human faces (\eg~FFHQ~\cite{karras2019style} and MetFaces~\cite{karras2020ada}) and bodies (\eg~SHHQ~\cite{fu2022stylegan} and DeepFashion~\cite{liu2016deepfashion}), scenes with a single salient subject (\eg~LSUN~\cite{yu2015lsun} and CompCars~\cite{yang2015large}), or multiple salient subjects (\eg~CLEVR\textit{n}~\cite{nguyen2020blockgan}). 
Most 3D-aware generative models handle face dataset and they are typically applicable to other dataset without any constraint, but there are other categories of 3D-aware generative model which need inductive bias of specialized domain knowledge.
Notably, recent research efforts~\cite{skorokhodov20233d,sargent2023vq3d} have been made to extend deep generative 3D-aware image synthesis beyond a single category, aiming to encompass a wide range of categories, such as those found in ImageNet~\cite{deng2009imagenet}. 
Open-source tools, such as Self-Distilled StyleGAN~\cite{mokady2022self}, are frequently utilized to facilitate the process of data generation. Examples of this include SDIP Dogs and SDIP Elephants.
}

\revise{\cref{tab:singledata} demonstrates a summary of single-view image datasets organized by their major categories and roughly sorted by their popularity how often they are used in studies. 
The following part provides a succinct overview of the commonly employed datasets in this field of research.
}

\vspace{1mm}
\revise{\textbf{SHHQ} (Stylish-Humans-HQ Dataset)~\cite{fu2022stylegan} , which was recently released, caters to the growing research interest in human body generation. It includes 230K high-quality, real-world, full-body human images, with resolutions ranging from 1024 $\times$ 512 up to 2240 $\times$ 1920. The SHHQ-1.0 subset, comprising 40K images, lays a solid foundation for extensive research and experimentation in the field of human body generation.}

\vspace{1mm}
\textbf{CelebA} (CelebFaces Attributes)~\cite{liu2015faceattributes} is a large-scale face attribute dataset consisting of 200K celebrity images with 40 attribute annotations each. CelebA, together with its succeeding CelebA-HQ~\cite{karras2017progressive}, are widely used in face image generation and manipulation.\par

\vspace{1mm}
\textbf{FFHQ} (Flickr-Faces-HQ)~\cite{karras2019style} is a high-quality image dataset of human faces crawled from Flickr, which consists of 70,000 high-quality human face images of 1024$\times$1024 pixels and contains considerable variation in terms of age, ethnicity, and image background.\par

\vspace{1mm} 
\textbf{AFHQ} (Animal-Faces-HQ)~\cite{karras2019style} consists of 70k high-quality animal face images of 512$\times$512 pixels. It includes three domains of cat, dog, and wildlife, each providing 5k images and containing diverse images of various breeds ($\geq$ eight). \par

\vspace{1mm} 
\textbf{CARLA}~\cite{dosovitskiy2017carla} is a synthetic dataset, which contains 10K images which are rendered from Carla Driving simulator~\cite{dosovitskiy2017carla} using 16 car models with different textures. \par

\vspace{1mm} 
\textbf{LSUN}~\cite{yu2015lsun} contains approximately one million labeled images for each of 10 scene categories (\eg, bedroom, church, or tower) and 20 object classes (\eg, bird, cat, or bus).
The church and bedroom scene images and car and bird object images are commonly used.\par

\vspace{1mm} 
\textbf{Megascans Food} (M-Food) and \textbf{Plants} (M-Plants) ~\cite{skorokhodov2022epigraf} are two \textit{variable-shape} datasets. They are proposed to address two limitations of existing simple-shape benchmarks: 1) they contain low variability of global object geometry, focusing entirely on a single class of objects, like human/cat faces or cars, that do not vary much from instance to instance; 2) they have limited camera pose distribution: for example, FFHQ and Cats are completely dominated by the frontal and near-frontal views. 
M-Food consists of 199 models of different food items with 128 views per model (25,472 images in total); and M-Plants consists of 1,108 different plant models with 128 views per model (141,824 images in total). 
Both contain images with $256\times 256$ pixels.

\vspace{1mm} 
\textbf{CLEVR\textit{n}}~\cite{nguyen2020blockgan} is a synthetic multiple-object dataset, where $n$ is the number of foreground objects. 
This dataset is derived from CLEVR~\cite{johnson2017clevr} by adding a large variety of colours and primitive shapes. 
In response to demand, the image can be rendered to a desired quantity (\eg~100k) and resolution (\eg~64$\times$64 in~\cite{nguyen2020blockgan} and 256$\times$256 in ~\cite{niemeyer2021giraffe}). 
This scheme can be used to generate other synthetic datasets, including~\textsc{SYNTH-CAR}\textit{n} and \textsc{SYNTH-CHAIR}\textit{n} with $n$ foreground objects each. 
These multiple-object datasets are often used to test models in terms of the independent control over foreground and background~\cite{nguyen2020blockgan,niemeyer2021giraffe}.

\revise{Besides single-view images, some studies~\cite{karnewar2023holodiffusion,anciukevivcius2023renderdiffusion} use multiview posed images for training. These involve synthetic data rendered from 3D scans, real images captured from the real world, or video frames from category-specific videos, along with corresponding camera parameters. For synthetic data, they use ground-truth camera poses, intrinsics, and bounds to render images from 3D shapes of objects (for example, from ScanNet~\cite{dai2017scannet} and ShapeNet~\cite{chang2015shapenet}). For real data (\eg~videos in the Common Objects in 3D (CO3D) dataset~\cite{reizenstein2021common}), off-the-shelf software, such as a structure-from-motion package like COLMAP~\cite{schonberger2016structure}, can be used to estimate the camera parameters.}

\subsection{Evaluation Metrics}
There are different dimensions to evaluate 3D-aware image synthesis methods, which can be categorized into two groups: 2D and 3D metrics. 
2D metrics evaluate the synthesised images in terms of quality, diversity, and fidelity. 
3D metrics access the shape and surface quality, as well as the  temporal and multi-view consistency.
Model efficiency is taken into account sometimes and is evaluated by model size and training/inference time.

\revise{The following part provides a succinct overview of the commonly employed measures.
It should be noted that, in the current evaluation landscape, no single metric can comprehensively address all aspects; instead, each typically examines a unique facet. Established metrics (despite their limitations) such as FID~\cite{heusel2017gans} and LPIPS~\cite{zhang2018unreasonable}, which are widely recognized within the research community, are frequently used to assess quality and diversity of generated images. On the other hand, the evaluation of 3D consistencies is less standardized, with various studies proposing their own metrics due to the lack of universally accepted 3D measures in this field.
Therefore, in addition to utilizing existing evaluation metrics, the introduction of more reliable and more personalized measures could significantly enhance the assessment of photorealistic and geometric quality of generated images in both general and specific contexts.
}

\subsubsection{Model Efficiency}

Two kinds of metrics are commonly employed by current studies to demonstrate the efficiency of their proposed methods: average running time and model complexity. 
These metrics can be borrowed from deep compression, which evaluates inference runtime, model size and latency. 
The model complexity is typically assessed by the number of parameters, floating point operations (FLOPs), and multiply-accumulate operations (MACs).
Runtime usually means one forward at inference phase. 
In our case, inference runtime means the time required for rendering an image. 
However, for deep 3D-aware generative models, we are equally interested in the time required for training. This can be assessed using metrics such as total training time and the number of batches processed per second~\cite{zhou2021cips3d}.

\tabsingledataset

\subsubsection{\revise{Image Similarity}}
\label{sec:faithfulness}

\revise{Similarity (or faithfulness) measures the similarity between real images and generated ones. 
When ground-truth images exist at certain viewpoints, the rendered images are expected to be close to them.
The most widely used metrics are Peak Signal-to-Noise Ratio (PSNR), Structural Similarity (SSIM)~\cite{TIP2004ImageWang}, and Learned Perceptual Image Patch Similarity (LPIPS)~\cite{zhang2018unreasonable}.}
Pixel-wise reconstruction distances, \eg~mean absolute error, are also used. %

\vspace{1mm} 
\textbf{PSNR} between the ground-truth image and the reconstruction is defined by the maximum possible pixel value of the image and the mean squared error between images.

\vspace{1mm} 
\textbf{SSIM} measures the structural similarity between images based on independent comparisons in terms of luminance, contrast, and structures.
The details can be found in~\cite{TIP2004ImageWang}.

\vspace{1mm} 
\revise{\textbf{LPIPS} measures the distance between image patches. A lower value means higher similarity between the image patches. A higher value means more different. 
LPIPS can therefore also be employed to assess the diversity of images.
The most common method for evaluating the diversity of generated images involves randomly selecting two examples from the set of generated images and calculating the average LPIPS distance between them. This measures the degree to which the two images differ from each other, with a higher LPIPS score indicating greater diversity.
} 

\subsubsection{Image Quality}
These metrics are often used to assess images generated by a generative model, like a generative adversarial network (GAN) or a generative diffusion model (GDM).
\revise{These metrics include but are not limited to Inception Score (IS)~\cite{salimans2016improved}, Fr$\acute{e}$chet Inception Distance (FID)~\cite{heusel2017gans}, Kernel Inception Distance (KID)~\cite{binkowski2018demystifying}, and Perceptual Path Length (PPL)~\cite{karras2019style}. 
At present, FID is the predominant metric in the research community for assessing the quality of generated images. In contrast, IS was once popular but has since been less favored due to its inability to accurately reflect the diversity of images and its sensitivity to minor changes in the image set. PPL measures the disentanglement and consistency of the learned latent space in generative models.}

\vspace{1mm} 
\textbf{FID} is defined by the Fr$\acute{e}$chet distance between features from the real and generated images based on Inception-v3~\cite{szegedy2016rethinking}.
Lower FID indicates better perceptual quality.

\vspace{1mm} 
\revise{\textbf{PPL} 
computes the average distance in feature space between images generated from interpolated points in the latent space. A smaller PPL suggests a smoother and more disentangled latent space, where minor changes in the latent vector lead to coherent and minor changes in the output image.}

\vspace{1mm} 
\textbf{KID} measures the dissimilarity between two probability distributions using samples drawn independently from each distribution. Lower is better.

\vspace{1mm} 
\textbf{IS} is to measure the quality and diversity of images generated from GAN models. 
It calculates the statistics of a synthesized image using Inception-v3 Network~\cite{szegedy2016rethinking} pretrained on ImageNet~\cite{deng2009imagenet}. 
A higher score is better.

\subsubsection{Multi-view 3D Consistency}
\label{subsubsec:multi-view_3D_consistency}

Multi-view 3D consistency is another significantly important aspect in 3D-aware image synthesis. 
The view-inconsistencies could be caused by shapes and colors. 
Consistencies in geometry and photometry is basically equivalent to the quality of shape and texture. 

\vspace{1mm} 
\textbf{Shape Quality} is evaluated mostly by calculating differences between the rendered depth map and the pseudo-ground-truth depth, \eg~using MSE~\cite{chan2022efficient} or a modified Chamfer distance~\cite{orel2022stylesdf}.
For example, given two generated images from two sampled angles of the same scene, Shi~\etal~\cite{shi20223d} uses rotation precision and rotation consistency to evaluate the quality of the depth maps (point cloud).  
The former is aimed to measure the accuracy of the angle of rotation while the latter targets at the rotation consistency evaluation.
In GOF~\cite{xu2021generative}, the mean angle deviation (MAD) and the scale-invariant depth error (SIDE) are used to compare the outputs against the ground-truth depth maps. MAD emphasizes the compactness of surfaces, whereas SIDE emphasizes the accuracy of depths.
Some methods use more direct indicators to evaluate the geometry properties of learned surfaces. Xu~\etal~\cite{xu2021generative} use average geodesic distance and average curvature between random points to assess the geometry properties of learned surfaces. The lower these two metrics, the smoother the recovered object surfaces. 

\vspace{1mm} 
\textbf{Texture Quality}  could be evaluated by using PSNR and SSIM as image fidelity under different viewpoints. But in most cases, the ground-truth images are not available for evaluation.
Several methods use FID to evaluate image quality at different camera poses as part of the multi-view texture quality.
A more direct way is to assess multi-view facial identity consistency (ID)~\cite{chan2022efficient} by calculating the mean Arcface~\cite{deng2019arcface} cosine similarity score between pairs of the same face rendered from random camera poses.

Apart from the aforementioned performance measures, pose accuracy~\cite{chan2022efficient} is also considered as an important indicator of shape quality and controllability. 
Poses (pitch, yaw, and roll) are detected with the help of pre-trained face reconstruction models from the generated images and then its L2 errors against the ground-truth poses is computed to determine each model’s pose drift.

\section{Overview of Deep Generative 3D-aware Image Synthesis}

\revise{In the following sections, we introduce different kinds of deep generative 3D-aware image synthesis methods. 
In this survey, we use the terms "generative 3D-aware image synthesis methods" and "3D-aware generative models" interchangeably, since they both fundamentally refer to the same concept related to image synthesis. \cref{tab:allcomparison} is a detailed comparison of deep generative 3D-aware image synthesis methods. The first (arXiv) draft dates are used to sequence the publications. The publication information can be found in the bibliography.}

\subsection{Goal and Challenge}

\revise{To begin with, we provide an overview of deep generative 3D-aware image synthesis to give readers a general understanding of this task, including its goals, challenges, and underlying principles. This serves to prevent readers from becoming overwhelmed with intricate technical details. As previously stated, the task involves learning a model capable of generating images that maintain consistency across multiple viewpoints, without explicitly modeling the subject(s) in 3D.
This definition itself implies goals and challenges of the task. The primary goal is to generate 3D-consistent images, while challenges lie in achieving multi-view consistency without explicit 3D shape modeling and solely relying on training datasets comprising unlabeled images without 3D supervision. Several potential issues arise consequently: 
\begin{itemize}
    \item Multiview inconsistency: One common issue is the presence of inconsistent shapes and appearances, where the textures and geometry vary across different views. Another challenge is background sticking, where the foreground subject is not adequately separated from the background.
    \item Limited camera views: The range of viewpoints provided by current models is often limited, with a lack of capability to generate images from less common perspectives (\eg~looking down from above or from behind). Furthermore, these models may not have the capability for 360-degree image generation.
    \item Imprecise camera control: Some models struggle to learn fine-grained control over camera parameters, leading to imprecise changes in the obtained images when altering the camera.
    \item Compromised visual quality: The image quality produced by 3D-aware generative models may not be as high as their 2D counterparts. Additionally, these models often generate images at lower resolutions, typically 256$\times$256 pixels, and rarely exceed resolutions of 1024 pixels.
    \item Limited application scenarios: Many current methods primarily concentrate on single categories characterized by simple geometry and appearance, such as human faces. These approaches demonstrate limited capabilities when it comes to composing complex scenes that encompass a variety of objects.
    \item Expensive training: Training 3D-aware generative models can be resource-intensive and time-consuming, requiring substantial computational power and extensive training time.
\end{itemize}
Despite significant advancements, these challenges continue to persist in this field. Most methods tend to tackle one or a few aspects, achieving satisfactory performance in those areas while leaving others compromised. The datasets and evaluation metrics introduced previously align with efforts to address these issues.}

\subsection{Comparisons between Two Primary Categories}

\revise{In this survey, deep generative 3D-aware image synthesis methods are classified into two main categories: 3D control of 2D generative models (\cref{sec:3d_control_of_2d_generative_models}) and 3D-aware generative models from image collections (\cref{sec:3d_generative_models}).
Then, every category is further divided into some subcategories depending on the experimental setting or the specific utilization of 3D information.
In particular, 3D control of 2D generative models are further divided into 1) 3D control in 2D latent spaces, 2) 3D parameters as controls, and 3) 3D priors as constraints.
For the category of 3D-aware generative models, most research relied on Generative Adversarial Networks (GANs)~\cite{goodfellow2020generative}, thus resulting in a prevalent exploration of 3D-aware GANs~\cite{nguyen2019hologan,nguyen2020blockgan,xue2022giraffehd,xu2022volumegan,gu2022stylenerf,orel2022stylesdf} (see~\cref{subsec:unconditional_3d_gans}). 
Recent trends in the studies are also indicating a growing interest in using diffusion models for 3D generative modeling~\cite{zhang20233dshape2vecset,kim2023neuralfield,muller2023diffrf,anciukevivcius2023renderdiffusion,karnewar2023holodiffusion,xiang20233d,chan2023generative} (see~\cref{subsec:unconditional_3d_diffusion}). 
The conditional 3D-aware generative models~\cite{sun2021fenerf,muller2022autorf,lin20223d,sun2022ide3d,cai2022pix2nerf,chen2022sem2nerf} are presented in~\cref{subsec:conditional_3d_generative_models}.
}

\revise{Typically, methods in the category of 3D control of 2D generative models exhibit superior image quality and require significantly fewer training resources. In contrast, methods in the second category can generate more consistent multiview images under a larger range of camera movements but with compromised visual quality, and training such models can be resource-intensive and time-consuming.
Therefore, improvements for methods in the first category primarily focus on enhancing multiview consistency, while the second category emphasizes the development of efficient and effective representations and rendering processes to improve visual quality and expedite training.
} 

\revise{Please note that as this field is relatively nascent, the usage of terminology can be perplexing and might lead to certain confusions or misunderstandings.
In certain literature, the term "3D-aware generative model" refers to methods discussed in~\cref{sec:3d_control_of_2d_generative_models}, while "3D generative models" are used to describe methods in~\cref{sec:3d_generative_models}.  
Given that 3D generative models are more aptly suited to 3D tasks, in the context of this survey, we categorize the \textit{methods discussed in \cref{sec:3d_control_of_2d_generative_models} as those based on a 2D network}, but which introduce various strategies to generate images consistent across multiple views. We refer to the \textit{methods in \cref{sec:3d_generative_models} as those utilizing a 3D-aware network design}.
We invite readers to focus on understanding the fundamental differences, rather than getting caught up in the specifics of terminology, and to make their own judgments based on the context.
}

\subsection{Relationships with INR-based Novel View Synthesis}
\label{subsec:relationship_with_inr_nvs}
 
\revise{This task of deep generative 3D-aware image synthesis, indicated by its nature, is closely related to 3D novel view synthesis (NVS)~\cite{sitzmann2019deepvoxels,sitzmann2019scene,sitzmann2021light,sajjadi2022scene,mildenhall2020nerf}, particularly the INR methods (\eg~NeRF~\cite{mildenhall2020nerf}). 
Many 3D-aware generative models (see~\cref{sec:3d_generative_models}) draw inspiration from the INR-based NVS methods~\cite{mildenhall2020nerf,martin2020nerf,yu2021pixelnerf,rematas2022urbannerf,gafni2021dynamic,muller2022instant,hu2022efficientnerf} to address aforementioned challenges.
Broadly speaking, both 3D-aware generative models and INR-based NVS methods aim to generate photorealistic and multi-view-consistent images, using a similar pipeline that first learns the implicit neural 3D representation and then renders it from that viewpoint. 
Particularly, both leverage neural 3D representations to represent scenes, use differentiable neural renderers to render them onto the image plane, and optimize the network parameters by minimizing the discrepancy between rendered and observed images.}

\revise{However, they are significantly different in training on a multiple-view or single-view image collections, due to their hugely different application scenarios. 
As with their 2D counterparts, 3D-aware generative models are learned from a collection of single-view images, while 3D novel view synthesis learns a 3D representation from multiple views of a scene.
Once trained, 3D-aware generative models generate images from a limited range of viewpoints, unlike the free-view rendering capabilities observed in 3D novel view synthesis (especially when utilizing NeRF-based methods). In contrast, NVS methods excel at synthesizing high-fidelity and detailed images from novel viewpoints, enabling the exploration of previously unseen perspectives.
}

\revise{Considering the intimate relationship between deep generative 3D-aware image synthesis and INR-based NVS, leveraging advancements made in INR-based NVS could potentially broaden the scope of camera movements in 3D-aware image generation. Furthermore, the combination of "free-view rendering and generative modeling" presents a challenging yet promising research avenue that is likely to gain significant attention in future research.
}

\section{3D Control of 2D Generative Models}
\label{sec:3d_control_of_2d_generative_models}

\revise{Due to the prevalence of 2D generative models, there have been studies aiming to make these pretrained models, especially GANs, 3D aware.
These studies, mostly built on the top of a pretrained StyleGAN, can be further categorized into three groups based on how the 3D control capability is introduced: 1) exploring 3D control in 2D latent spaces,
2) adopting explicit control over the 3D parameters, and 3) introducing 3D-aware components into 2D GANs.
The same taxonomy can be applied to other generative models.
\cref{fig:group_1_comparison} is an illustration of these three categories of methods.}

\taballcomparison

\subsection{\revise{Exploring 3D Control in 2D Latent Spaces}}
\label{subsec:exploring_3d_control_in_2d_latent_space}

\subsubsection{Discovering 3D Control Latent Directions}
\label{subsubsec:discovering_3D_control_latent_directions}

It has been demonstrated that pretrained GANs have interpretable directions in their latent spaces. The image generation process is controlled by altering the latent codes $\z$ in the desired directions $\n$ with step $\alpha$, which is often considered as a linear vector arithmetic $\z^{\prime}=\z+\alpha\n$.
The altered latent codes are then fed into a pretrained GAN $G(\cdot)$ for the edited results: $I^{\prime}=G(\z^{\prime})$.
The methods in this group are mostly developed for semantic editing, and some have been shown to discover geometric directions as well.
They are used to alter pose position or light condition of faces or manipulate geometry (\eg~zoom, shift, or rotation) of natural images.
As classified in~\cite{xia2022gan}, such directions can be identified through supervised, self-supervised, or unsupervised manners.

\vspace{1mm}
\noindent\textbf{Supervised Manner}
These methods typically sample a large amount of latent codes, synthesize a collection of corresponding images, and annotate them with predefined labels by introducing a pretrained classifier.
For example, to interpret the face representation learned by GANs, Shen~\etal~\cite{shen2020interpreting} (May 2020) employ some off-the-shelf classifiers to learn a hyperplane in the latent space serving as the separation boundary and predict semantic scores for synthesized images. 
Even though the boundary is searched by solving a bi-classification problem, it can produce continuous face pose changing by moving the latent code.
Abdal~\etal~\cite{abdal2020styleflow} (Aug 2020) learn a bidirectional mapping between the $\mathcal{Z}$ space and the $\mathcal{W}$ space by using continuous normalizing flows (CNF).  Attribute information (including head poses) are injected into the CNF blocks for the desired results.
However, such methods rely on the availability of attributes (typically obtained by a face classifier network), which might be difficult to obtain for new datasets and could require manual labeling effort.
Jahanian~\etal~\cite{jahanian2020steerability} (Jul 2019) use a self-supervised manner to learn these directions without any direct supervision. 
Sequence of target images are obtained by applying simple augmentations to the source image. 
Specifically, image shifting is used for camera motion along vertical and horizontal axes, downsampling and central cropping for zooming in and out, and perspective transformation for rotation.
With inverted images $G(\z)$ and target edits \texttt{edit}$(G(\z), \alpha)$, they learn the direction $\n$ %
by minimizing the distance between the generated image $G(\z+\alpha \n)$ after taking an $\alpha$-step in the latent direction and the target image \texttt{edit}($G(\z), \alpha$). 

\vspace{1mm}
\noindent\textbf{Unsupervised Manner}
Some methods~\cite{voynov2020latent,eric2020GANSpace} aim to discover interpretable directions in the latent space in an unsupervised manner, \ie, without the requirement of paired data.
For example, Härkönen~\etal~\cite{eric2020GANSpace} (Apr 2020) create interpretable controls for image synthesis by identifying important latent directions based on PCA applied in the latent or feature space. The obtained principal components correspond to certain attributes, and the selective application of the principal components allows for the control of many image attributes. 
Some of their discovered directions support 3D operations such as rotation or zooming out.
This method is considered as ``unsupervised'' since the directions can be discovered by PCA without using any labels.
There is still a need to manually annotate these directions to the target operations and to which layers they should be applied to.

In~\cite{shen2021closedform}, Shen~\etal~(Jul 2020) show that latent directions in a pretrained GAN for 3D-aware image synthesis can be directly computed in a closed form without any kind of training or optimization. 
They propose a \textbf{Se}mantics \textbf{Fa}ctorization (SeFa) method based on the singular value decomposition of the weights of the first layer of a pretrained GAN.
They observe that the semantic transformation of an image, usually denoted by moving the latent code toward a certain direction $\n^{\prime} = \z + \alpha \n$, is only determined by the latent direction $\n$.
Therefore, the directions $\n$ can cause a significant change in the output image $\Delta\y$, \ie, $\Delta\y =\y^{\prime}-\y= (\A(\z + \alpha\n) + \bb) - (\A\z + \bb) = \alpha\A\n$, where $\A$ and $\bb$ are respectively the weight and bias of certain layers in $G$. 
The obtained formula, $\Delta\y = \alpha\A\n$, suggests that the desired editing with direction $\n$ can be achieved by adding the term $\alpha\A\n$ onto the projected code and indicates that the weight parameter $\A$ should contain the essential knowledge of image variations.
The eigenvectors of the matrix $\A^T\A$ should be the desired directions $\n^*$. This gives a closed-form factorization of latent semantics in GANs.
This method supports multiple 3D control operations such as car orientation, face and body pose, streetscape and bedroom viewpoint, zoom, shift, as well as rotation on a variety of GANs.

\revise{%
While previous methods predominantly relied on GANs, diffusion models have recently emerged as compelling alternatives. Despite their increasing popularity in image synthesis and editing, the understanding of their latent space is still under exploration.
Recently, Kwon~\etal~\cite{kwon2022diffusion} introduce an Asymmetric Reverse Process (Asyrp) strategy to discover a semantic latent space in frozen pre-trained diffusion models. They coin this semantic latent space for Denoising Diffusion Models (DDMs) as "h-space", which has shown its potential in facilitating semantic image editing in a manner akin to GANs. This h-space consists of the bottleneck activations in the DDM's denoiser at each timestep of the diffusion process.
Building upon this discovery, Haas~\etal~\cite{haas2023discovering} delve deeper into understanding the properties of h-space, proposing several innovative methods to discover meaningful semantic directions within it.
In a different approach, Brack~\etal~\cite{brack2022stable} put forward the Stable Artist, designed to guide the semantic direction in the latent space of text-conditioned generative diffusion models. Their primary component, Semantic Guidance (SEGA), steers the diffusion process along variable numbers of semantic directions. This offers the ability to make subtle image edits, alter compositions and styles, as well as optimizing overall artistic conception. Beyond these capabilities, SEGA also enables probing of latent spaces to uncover insights into how the model represents learned concepts.
}

\subsubsection{\revise{Pinpointing Predetermined Targets}}
\label{subsubsec:pinpointing_predetermined_destination}

\revise{Similar to discovering semantic directions, recent methods~\cite{pan2023drag,endo2022user,mou2023dragondiffusion,shi2023dragdiffusion} have proposed alternative strategies to pinpoint desired edits in the latent space of a pretrained generative model. Unlike the predetermined route provided by the discovered semantic directions, these methods present a different outlook: they acknowledge the goal (or destination) but do not necessitate knowing the specific path to reach it. Such methods are typically formulated as optimization problems, aiming to determine the alterable trajectory from a starting point to the predetermined destination~\cite{xia2021tedigan,xia2021open}.
For instance, the ability to manipulate coarse object position is realized by integrating intermediate constructs, such as "blobs"~\cite{epstein2022blobgan}, or heatmaps~\cite{wang2022improving}. These strategies facilitate the modification of either image-aligned semantic attributes, like appearance, or broad geometric characteristics, including object position and pose.
In contrast to these methods, a few methods utilize point-based editing, a powerful but less-explored way of controlling GANs.
In GANWarping~\cite{wang2022rewriting}, the user is asked to warp a select number of generated images by defining several control points to create customized models. 
Although the modifications alter the shape of the object, other visual elements such as pose, color, texture, and background are faithfully maintained. However, the realism of the warped images is not ensured.
UserControllableLT~\cite{endo2022user} allows point-based editing by transforming the latent vectors of a GAN. However, this approach only supports editing using a single point being dragged on the image, and it does not handle multiple-point constraints effectively. Additionally, the control is not precise; the target point is often not reached after editing.
DragGAN~\cite{pan2023drag}, on the other hand, enables users to "drag" any points of an image to precisely reach target points interactively. 
DragGAN iteratively performs motion supervision and point tracking. The motion supervision directs the handle point towards target position; the point tracking updates the handle point to track the object in the image.
Through DragGAN, users can deform an image with precise control over pixel placement, thereby manipulating pose, shape, expression, and layout of various categories such as animals, cars, humans, landscapes, and more.}
\revise{Inspired by the advancements in dragging GANs~\cite{endo2022user,pan2023drag}, a handful of studies~\cite{mou2023dragondiffusion,shi2023dragdiffusion} have begun to explore similar drag-style manipulations on Diffusion models.
}

\subsection{Incorporating 3D Parameters as Controls}
\label{subsec:3d_parameter_as_controls}
Methods using 3D parameters as control factors typically follow a paradigm described as $x^{\prime} = G(x, \theta)$.
Explicit control over the 3D parameters $\theta$ gives the edited result $x^{\prime}$. 
Here, $\theta$ could be human-interpretable attribute descriptions or a set of parameters from 3D models. 
Sometimes, the given control factors are intuitively understandable, \eg, (age: 20 years old), (head pose: pitch, yaw, roll), or using the environment map to represent light condition.
Most methods in this category incorporate 3D pretrained model parameters into 2D image-based generative models for controllable 3D-aware synthesis.
These methods propose solutions to translate controls of 3D face rendering models into GAN-generated processes.
Taking face generation as an example, they usually integrate priors from a parametric 3D Morphable Model (3DMM)~\cite{blanz1999morphable} as explicit control factors.
\cref{tab:explicitcontrol} is an overview of methods that incorporate 3D parameters to a 2D generative model.

\figcomparison

The models in this section, as well as those in the next, make use of additional 3D models. Their main difference is that the former uses 3D model parameters as input control factors while the latter uses them as supervision signals.
There is another series of studies combining implicit 3D representation with 3DMM, either being trained with a reconstruction loss using annotated multi-view datasets~\cite{hong2022headnerf} or directly imposing 3DMM conditions into 3D NeRF volume and being trained on unannotated single-view images~\cite{sun2022controllable}.
We focus on leveraging 3D priors for image synthesis based on 2D generative models and will introduce other studies in the remaining sections.

\subsubsection{\comments{Preliminary:} Control Factors from Pretrained Models} \label{subsubsec:control_factors}

Most methods use 3DMM parameters to provide explicit control. 
3DMMs is commonly used to represent faces, where faces are parameterized by head rotation $\phi$ and translation $\rho$, identity geometry $\alpha$, expressions $\beta$, skin reflectance $\delta$, and scene illumination $\gamma$: $\theta=(\phi, \rho, \alpha, \delta, \beta, \gamma) \in \mathbb{R}^{m}$.
The parametric nature of 3DMMs allows navigating and exploring the space of plausible faces.
Thus, synthetic images can be rendered based on different parameter configurations.
In practice, these 3DMM parameters are first transformed before being used in the network~\cite{tewari2020pie,deng2020disentangled}. 
Besides 3DMM, parameters from other state-of-the-art tools could also be used to provide 3D control factors. 
In~\cite{abdal2020styleflow}, Microsoft Face API predicts pitch and yaw as the head pose. 
GAN-Control~\cite{shoshan2021gan} extracts head-pose, expression, illumination, age, and hair color by using several off-the-shelf attribute predictors.
DiscoFaceGAN~\cite{deng2020disentangled} extracts identity, expression, and texture information from 3DMM, approximates scene illumination with Spherical Harmonics (SH)~\cite{ramamoorthi2001efficient}, and defines face pose as three rotation angles.

For this category, a key question is how to associate these parameters with corresponding images as these methods require supervised training. 
Except one using existing synthetic data with 3D parameters~\cite{kowalski2020config}, others use pretrained models to achieve transformations from one direction to another, \ie~$I \rightarrow \theta$ by using attribute predictors~\cite{shoshan2021gan} or $\theta \rightarrow I$ by synthetic rendering~\cite{tewari2020stylerig,liu20223d}, and learn a mapping network for the opposite direction.

\subsubsection{Explicit Control over 3D Parameters}
\label{subsubsec:explicit_control_over_3d_parameters}

\revise{With 3D parameters obtained, many methods~\cite{tewari2017mofa,tewari2020stylerig,tewari2020pie,liu20223d,shoshan2021gan,kowalski2020config} are developed to incorporate them as input control factors into a 2D generative model for controllable 3D-aware image synthesis.}
This section demonstrates how these methods introduce 3D parameters and achieve explicit control through data collection, network design, and loss functions.

In MoFA~\cite{tewari2017mofa}, Tewari~\etal~(Mar 2017) use a CNN to project a face into the 3DMM space, followed by a differentiable renderer to synthesize the reconstructed face. 
The network is trained on a large collection of face images in a self-supervised manner.
Inspired by the computer graphics pipeline, CONFIG~\cite{kowalski2020config} (May 2020) uses a set of parameters to represent and control desired factors.
Blendshape values control facial expressions, Euler angles control head pose, and environment maps control the illumination.
CONFIG has two encoders ($E_R$ and $E_S$) that encode real face images $I_R$ and the parameters $\theta: \{\theta_1,\cdots,\theta_k\}$ of the synthetic images to a shared latent space $\mathcal{Z}$, which is factorised into elements that each part $z_i$ corresponds to a different facial attribute controlled by $\theta_i$. 
Each element $z_i$ comes as the $i$-th parameter of $z \in \mathcal{Z}$ either from $\theta_i$ (encoded by $E_S$) or from a different real face image (encoded by $E_R$).
They adopt a two stage-training scheme to learn a disentangled latent space and produce photorealistic images.

Those based on StyleGANs~\cite{karras2019style,karras2020analyzing} either use a pretrained StyleGAN model that keeps its weights fixed~\cite{tewari2020stylerig,tewari2020pie} or make slight modifications to how 3D parameters are incorporated as conditions~\cite{shoshan2021gan,deng2020disentangled}.
StyleRig~\cite{tewari2020stylerig} and PIE~\cite{tewari2020pie} are two examples of using a pretrained StyleGAN.
StyleRig (Apr 2020) trains a neural network, called RigNet, to inject a subset of parameters into a given StyleGAN latent code $w$.
RigNet is a function $\text{rignet}(\cdot, \cdot) $ that maps a pair of StyleGAN code $w$ and subset of 3DMM parameters $\theta$ to a new StyleGAN code $w^{\prime}$, \ie~$w^{\prime} = \text{rignet}(w, \theta)$. 
Several RigNets are trained, each dealing with a single mode of control (pose, expression, lighting). 
For self-supervised training, they introduce two key components: a learnable parameter regressor $\mathcal{F}$ and a pretrained differentiable render layer $\mathcal{R}$.
$\mathcal{F}$ maps a latent code $w$ to a vector of semantic control parameters $\theta$: $\theta=\mathcal{F}(w)$.
$\mathcal{R}$ takes a parameter vector $\theta$ as input and generates a synthetic rendering $I_w = \mathcal{R}(\theta)$.
StyleRig allows for multiple-attribute editing but only on synthetic facial images rather than real ones.
In contract, PIE (Sep 2020) uses a model-based face auto-encoder to replace $\mathcal{F}$ and $\mathcal{R}$ of StyleRig in support of real image editing.

Some methods inherit the main structure of Style-based generators and make slight modifications, mainly different on how the 3D parameters are incorporated as the condition.
DiscoFaceGAN~\cite{deng2020disentangled} (Apr 2020) proposes an unconditional 3D-aware method with controllability on four attributes: identity $\alpha$, expression $\beta$, scene illumination $\gamma$, and face pose $\delta$.
Their model consists of two networks that learn the mapping 1) $V(\cdot)$ from $z$-space to $\theta$-space; and 2) $G(\cdot)$ from $\theta$-space to the image space.
The latent code $z$ is sampled from a standard normal distribution. 
The parameters $\theta$ is the concatenation of the four control factors $\theta :=[\alpha, \beta, \gamma, \delta, \varepsilon]$ and the noise $\varepsilon$, which is the same of $z$ for image diversity. 
They train four VAEs for $\alpha$, $\beta$, $\gamma$, and $\delta$ on the $\theta$ samples extracted by using a off-the-shelf 3D face reconstruction method from real image set.
Only the decoders are kept after the VAE training and denoted as $V_i, i= 1, 2, 3, 4$, for z-space to $\theta$-space mapping.
For training $G$, they sample $z = [z_1,\cdots, z_5]$ from standard normal distribution, map it to $\theta$, and feed $\theta$ to both $G$ and the renderer to obtain a generated face $x$ and a rendered face $x^\prime$, respectively.
They apply three types of losses for training: adversarial loss, imitative loss, and contrastive loss.
GAN-Control~\cite{shoshan2021gan} (Jan 2021) builds on the StyleGAN2~\cite{karras2020analyzing} architecture. 
They divide the $\mathcal{Z}$ and $\mathcal{W}$ latent spaces to $N+1$ separate sub-spaces, in accordance with $N$ control attributes and one residual one for non-concerned information.
The original StyleGAN2 architecture is changed from a single mapping network ($w=f(z)$, implemented as an eight-layered MLP) to each control $z_i$ having its own $f(\cdot)$ so that $w_i=f(z_i)$. 
The combined latent vector (concatenation of the sub-vectors), $w$, is then fed into the generator $G$.
To enable explicit control over each attribute, they use contrastive learning for disentanglement.
Given a set of pretrained attribute predictors $\{\mathcal{R}_i\}_{i=1}^N$, they extract intermediate features as the attribute information from sampled images $G(z)$ and use them to calculate the distances during training.
To support explicit control during inference, they further train $N$ encoders $\{E_k\}_{k=1}^N$, each to map a human-interpretable attribute representation $y^k$ to a latent code $w^k$.
They use the attribute predictors to label the randomly-sampled images as the training data.
Different from GAN-Control~\cite{shoshan2021gan}, in 3D-FM GAN~\cite{liu20223d}, Liu~\etal~(Aug 2022) change StyleGAN $G$ to make it conditional on a given image and a rendering. They estimate the lighting and 3DMM parameters of the face as the 3D parameters $\theta$. These $\theta$ are not directly incorprated into $G$ as the explict control signal but are used instead to generate a rendering $I_r(\theta)$ of the same given image $I(\theta)$, which leads to a paired dataset.
The resulting pairs $\{I(\theta), I_r(\theta)\}$ are used for reconstruction training, and $\{I({\theta}_i), I_r({\theta}_j)\}$ with different attributes of the same identity are for disentangled training.

\tabexplicitcontrol

\subsection{Introducing 3D prior knowledge as Constraints}
\label{subsec:3d_prior_as_constraints}

This category of studies facilitate the learning of 3D consistency by utilizing one or more kinds of 3D prior knowledge as constraints, such as shape~\cite{wu2016learning,zhu2018visual,chen2021towards}, albedo~\cite{abu2018geometric,shi2021lifting}, normal~\cite{abu2018geometric}, and depth~\cite{noguchi2020rgbd,shi2021lifting,shi20223d}.
Both~\cref{subsec:3d_parameter_as_controls} and this section aim to make a 2D generative model, especially GAN, 3D-aware.
\cref{subsec:3d_parameter_as_controls} focuses on the methods that incorporate 3D prior knowledge to 2D GANs for explicit control, while this section emphasizes those methods that introduce 3D-aware components into 2D GANs and use 3D prior knowledge as constraints for training.
In addition to whether to introduce explicit 3D parameters as inputs, the slight difference between them is also reflected in the different concepts of dataset usage and network structure design.
The former is able to control each of the desired attributes because of introducing 3D parameters as control factors but it lacks explicit geometry and texture as holistic 3D supervision, which leads to multi-view inconsistencies.
The latter introduces 3D-aware components into 2D generative models (mostly GANs) and uses 3D prior knowledge (\eg~predicted depth from a off-the-shelf depth estimation method) to constrain the training process, resulting in a degree of consistency but in the meantime a lack of fine-grained control.
The two types of methods are not mutually exclusive. In addition to introducing 3D parameters to improve controllability, a few methods also implement 3D constraints to improve consistency across multiple views.

\subsubsection{\comments{Preliminary: }3D Prior Knowledge}
\label{subsubsec:3d_prior_knowledge}

Basically, the intrinsic components used to describe the physical world can be used here as priors, including but not limited to shape (surface, depth, and normal), material (albedo, reflectivity and shininess), and lighting (direction, intensity). 
We introduce in~\cref{subsec:3d_shape_representation} some common shape representations.
Albedo, also referred to as reflection coefficient, is a measure of how reflective a surface is. It is either determined by a value between 0 and 1 or a percentage value. The more reflective a surface is, the higher the albedo value.
A surface normal, or simply normal, to a surface at each point is a vector perpendicular to the tangent plane of the surface at that point.
Normals represent the curvature of the object and can be used for reflecting light.
Depth is the distance between the camera and the object at each pixel.
They all contain geometric contextual features.
There is a track of studies of intrinsic decomposition, which can be seen as a simplification of inverse rendering for general scenes, aiming to provide interpretable intermediate representations from images.
There are also many methods specifically proposed to infer one specific environmental component, such as depth estimation, normal estimation, and light estimation. 
All these models can be potentially used as the training constraints for this category of methods.

\figdcomparison

\subsubsection{Introducing 3D Components into 2D Models}
\label{subsubsec:introducing_3d_components_into_2d_gans}
With the chosen 3D priors, the next important decision for this kind of methods is to find a way of introducing 3D-aware components into their models and using 3D priors as training constraints.
In S$^2$-GAN~\cite{wang2016generative}, Wang~\etal~ (Mar 2016) use a two-stage training for indoor scene synthesis: an unconditional GAN for structure (geometry) generates a surface normal map and the second GAN for style (appearance) takes this surface normal map as condition and outputs an image.
VON~\cite{zhu2018visual} (Dec 2018) uses shape as the 3D prior and design a GAN based on 3D convolutional neural network to learn the geometry information. 
An unconditional shape GAN $G_s$ first generates voxel grid shape $s$ from a randomly sampled shape code $z_s$. The differentiable projection module then projects $s$ to 2.5D sketches $s_{2.5D}$ at a sampled viewpoint $z_v$. 
The 2.5D sketches include both the object’s depth and silhouette.
The texture network $G_t$ finally adds realistic, diverse texture to these 2.5D sketches to generate 2D images: $G_t$ takes $s_{2.5D}$ and another latent code $z_t$ as input and outputs a 2D images.
GIS~\cite{abu2018geometric} (Sep 2018) and NGP~\cite{chen2021towards} (Feb 2021) utilize more than one 3D prior, such as albedo maps and normal maps, resulting in multiple 2D GANs to learn all the 3D attributes. 
In the above methods, 3D-aware components are used as intermediates, which are supervised by outputs from pretrained models.
There are a few methods that only introduce 3D-aware components into 2D models without using 3D priors from pretrained models to constrain the training. 
For example, RGBD-GAN~\cite{noguchi2020rgbd} (Sep 2019) generates two RGBD images with different camera parameters and then warps them to each other to ensure 3D consistency. 
It learns to generate view-consistent images consistent from pure 2D image collections.

More recently, a few methods~\cite{shi2021lifting,shi20223d} are build on top of StyleGAN architectures, either using a pretrained StyleGAN or adapting the vanilla design to their setting. They are also referred to as StyleGAN2-based 2.5D GANs in~\cite{chan2022efficient}.
LiftedGAN~\cite{shi2021lifting} (Nov 2020) equips a pre-trained StyleGAN2 generator with five additional 3D-aware networks, which disentangle the latent space of StyleGAN2 into texture, shape, viewpoint, and lighting. These 3D components are then used for rendering. The proposed model is able to output both the 3D shape and texture, allowing explicit pose and lighting control.
To control the viewpoint, DepthGAN~\cite{shi20223d} (Feb 2022) designs a dual-generator based on StyleGAN. The depth branch $G_d$ takes as the input an uniformly sampled angle $\theta$ from range $[\theta_l, \theta_r]$ and a depth latent code, and synthesize a depth image at $\theta$. The rgb branch $G_r$ takes the intermediate feature maps of $G_d$ as the conditions to acquire the geometry information. They use a pre-trained depth prediction model to get the corresponding depth image of each RGB image. A rotation consistency loss is introduced to enhance the multi-view consistency during training. The image synthesized under angle $\theta_1$ is projected to a point cloud and re-projected to the 2D space under $\theta_2$, and compared with the image generated under $\theta_2$.
\revise{GMPI~\cite{zhao2022generative} (Jul 2022) makes a classical 2D GAN, \ie~StyleGAN2, 3D-aware by only introducing 1) a multiplane image style generator branch which produces a set of alpha maps conditioned on their depth; and 2) a pose conditioned discriminator.
}

Despite impressive image quality, these methods still tend to produce 3D inconsistent faces under large expression and pose variations or scenes under different views due to the lack of a holistic 3D representation. 
They also inherit inconsistencies introduced by the pretrained model they use. For example, depth estimation methods, especially depth estimated from a single image, are known to suffer from the world-inconsistency. 
\revise{With the advances in differentiable rendering and implicit neural 3D representations, a recent line of work has explored photorealistic 3D-aware face or scene synthesis using~\textit{only} 2D image collections as the training data, \textit{without any 3D supervision}. 
The representative studies are categorized into INR-based NVS methods and 3D-aware generative models.
The details of INR-based NVS methods can be found in the appendix, while \cref{subsec:3d_novel_view_synthesis} and \cref{subsec:relationship_with_inr_nvs} provide a brief introduction and a thorough discussion of their relationships, respectively. We delve into 3D-aware generative models in the subsequent section.
}

\section{3D-aware Generative Models}
\label{sec:3d_generative_models}

Inspired by 3D novel view synthesis methods, follow-up works introduce the efficient and expressive neural scene representations, especially INR to the field of 2D generative image synthesis, leading to a new paradigm called \textit{3D-aware generative models}~\cite{chan2021pigan,schwarz2020graf}.
These methods do not assume a large number of posed images of a single scene. Instead, they learn a model for synthesizing novel scenes by training  on 
single-view images without 3D supervision.
\revise{As pointed out in~\cref{subsec:relationship_with_inr_nvs}, the INR-based novel view synthesis techniques and the methods discussed in this section share common terminology and objectives. As such, they frequently draw inspiration from one another.
Both aim to generate multi-view-consistent and photorealistic images, using a similar pipeline that first learns the 3D representation and then renders it from that viewpoint (see~\cref{fig:group_23_comparison}).
It is their application scenarios and training data that differentiate the two kinds of methods. As with their 2D counterparts, 3D-aware generative models generate images from a collection of single views, while 3D novel view synthesis learns a 3D representation from multiple views of a scene.
}

\revise{These 3D-aware generative models follow a similar experimental setting as their 2D counterparts, \ie, generating high quality photorealistic results from single-view image datasets, with an extra goal to ensure 3D consistency across multiple views. 
In 2D generative models, ``unconditional'' methods are referred to as those merely inputting latent codes that are sampled from a prior distribution.
In this survey, we use ``unconditional'' 3D-aware generative models to refer to those using latent codes and camera positions as input. In some cases, the camera positions could also be generated from the randomly-sampled latent codes instead of human-understandable control factors.
Those taking other inputs, especially image, text, semantic label, or sketch, are the conditional ones.}

\revise{Much of the earlier research primarily on unconditional generation relied on 3D-aware Generative Adversarial Networks (GANs), thus resulting in a prevalent exploration of 3D-aware GANs~\cite{nguyen2019hologan,nguyen2020blockgan,xue2022giraffehd,xu2022volumegan,gu2022stylenerf,orel2022stylesdf} (see~\cref{subsec:unconditional_3d_gans}). However, recent trends in the studies are also indicating a growing interest in using diffusion models for 3D generative modeling~\cite{zhang20233dshape2vecset,kim2023neuralfield,muller2023diffrf,anciukevivcius2023renderdiffusion,karnewar2023holodiffusion,xiang20233d,chan2023generative,ma2023adding} (see~\cref{subsec:unconditional_3d_diffusion}). 
}
\revise{We then present conditional 3D-aware generative models~\cite{sun2021fenerf,muller2022autorf,lin20223d,sun2022ide3d,cai2022pix2nerf,chen2022sem2nerf} in~\cref{subsec:conditional_3d_generative_models}.}

\subsection{Unconditional 3D-aware GANs}
\label{subsec:unconditional_3d_gans}

\revise{The majority of methods in unconditional 3D-aware generative models belong to the category of GANs. 3D GANs, an outstanding representative of 3D-aware generative models, usually utilize an adversarial framework to learn these representations in an unsupervised manner.
The performance details of several representative unconditional 3D-aware GANs~\cite{schwarz2020graf,chan2021pigan,orel2022stylesdf,gu2022stylenerf,xu2022volumegan,deng2022gram,skorokhodov2022epigraf,chan2022efficient} on the FFHQ dataset~\cite{karras2019style} are provided in~\cref{tab:quantitative_conditional_3d_gans}, with the FID~\cite{heusel2017gans} used as the evaluation metric.
Some use generative latent optimization~\cite{rebain2022lolnerf,bojanowski2017optimizing} instead of adversarial training~\cite{goodfellow2020generative}. 
Other kinds of generative models, especially diffusion models~\cite{ho2020denoising,song2020denoising}, which have proven extremely effective in generating high-quality images in recent years, have not yet been widely applied to 3D-ware image synthesis. 
Only very few recent studies are based on diffusion models for 3D novel view synthesis~\cite{watson2022novel}, which are introduced in \cref{subsec:unconditional_3d_diffusion}. 
These generative models are expected to catch up in the near future.
}

\tabquangans

Like 2D GANs, 3D-aware GANs have recently achieved tremendous breakthroughs in terms of image quality and editability for 2D image synthesis, with the extra goal of 3D consistency by introducing explicit camera control. These methods can be formulated in the form of $I = f(z, \theta)$, where the noise vector $z$ is for appearance and $\theta$ means camera pose.
Towards editable, high-resolution, and view-consistent image synthesis, proposed methods mainly work on two key components: 1) to learn efficient and expressive representations of geometry and appearance; 2) to develop accelerated and view-consistent rendering algorithms. Some methods are proposed to facilitate user-interactive editing. 

\revise{In this section, we first introduce the unconditional 3D-aware GANs based on different neural scene representations.
We especially emphasize their efforts towards: 1) learning efficient and expressive geometry and appearance representations (\cref{subsubsec:efficient_and_expressive_3d_representation}); 2) developing accelerated and view-consistent rendering algorithms (\cref{subsubsec:accelerated_and_consistent_rendering}); 3) broadening the applicable scenarios (\cref{subsubsec:broadening_applicable_scenarios}); and 4) real-time and user-interactive editing (\cref{subsubsec:interactive_editing}). 
}

\subsubsection{Efficient and Expressive Representations}
\label{subsubsec:efficient_and_expressive_3d_representation}

Early 3D GANs adopt voxel-based representation.
For example, PrGANs~\cite{gadelha20173d} (Dec 2016) and VON~\cite{zhu2018visual} (Dec 2018) first learn an explicit shape and render images at different viewpoints. They are trained under 3D supervision (as such they are also categorized into~\cref{subsec:3d_prior_as_constraints}).
PlatonicGAN~\cite{henzler2019escaping} (Nov 2018) learns a generative 3D model from an unstructured collection of 2D images. 
The learned shape and its rendered images are limited to low resolutions and coarse detail due to the computational complexity.
The approaches using deep-voxel representations~\cite{nguyen2019hologan,nguyen2020blockgan} can create finely-detailed images under different poses.
HoloGAN~\cite{nguyen2019hologan} (Apr 2019) first uses 3D convolutions to learn a deep-voxel representation (in a canonical pose), then utilizes a rigid-body transformation (3D rotation) to transform this representation to a certain pose, and finally applies a projection unit to render an image.
In contrast to HoloGAN, which learns 3D features directly for the entire scene, BlockGAN~\cite{nguyen2020blockgan} (Feb 2020) learns 3D features for each object separately.
It decomposes a 3D scene into a background and one or more foreground objects, each of which is represented by a noise vector $z_i$. 
This design disentangles a scene into separate objects and enables control over camera pose, lighting, and shadow.
However, early voxel-based methods~\cite{gadelha20173d,zhu2018visual,henzler2019escaping,nguyen2019hologan,nguyen2020blockgan} fail to synthesize complex scenes and photorealistic details due to the limited grid resolutions.
Their reliance on a learned black-box rendering leads to discretization artifacts, degrades view-consistency of the generated images, and makes generalization to unseen camera poses difficult.
Liao~\etal~\cite{liao2020towards} (Dec 2019) use 3D primitives as abstract object representations and differentiable rendering to project the 3D representations onto the image plane where a 2D generator transforms them into object appearances and composites them into a coherent image.

The limited expressiveness and efficiency of previous methods prevents them from synthesising complex scenes and photorealistic details.
Therefore, INRs, especially NeRFs, which have proven to generate high-fidelity results in novel view synthesis, are introduced to 3D-aware generative models. 
To avoid the requirement of posed images, an increasing number of methods
turn to utilize an adversarial framework to train a generative model for these representations from \textit{unposed} images.
\revise{The visualization results from COLMAP~\cite{schonberger2016structure} validate those methods showing greater 3D consistency than the voxel-based representations.}
GRAF~\cite{schwarz2020graf} (Jul 2020) uses NeRF to represent the scene and an adversarial framework to train on unposed images.
The generator takes camera matrix, camera pose, 2D sampling pattern, and shape/appearance codes as input and predicts an image patch. The discriminator compares the synthesized patch to a real patch extracted from a real image.
One significant modification is that GRAF makes NeRF conditioned on two additional latent codes: a shape noise and an appearance noise. 
A follow-up work pi-GAN~\cite{chan2021pigan} (Dec 2020) differs from GRAF in three ways on network architecture and training strategy: 
1) pi-GAN uses SIREN~\cite{sitzmann2020implicit} as the choice of scene representation rather than a positionally-encoded ReLU MLP~\cite{mildenhall2020nerf}; 
2) pi-GAN leverages a StyleGAN-inspired mapping network to condition layers in the SIREN on a single input noise code through feature-wise linear modulation (FiLM) instead of conditioning on two additional shape/appearance codes;
3) pi-GAN follows ProgressiveGAN~\cite{karras2017progressive} where discriminator grows progressively rather than a patch-based discriminator.
Built similarly to pi-GAN, LOLNeRF~\cite{rebain2022lolnerf} (Nov 2021), which is capable of single-shot view synthesis of human faces, uses generative latent optimization~\cite{bojanowski2017optimizing} instead of adversarial training~\cite{goodfellow2020generative}.
GIRAFFE~\cite{niemeyer2021giraffe} improves the BlockGAN~\cite{nguyen2020blockgan} framework by replacing the voxel-based representation and 3D-to-2D projection with a NeRF-based compositional 3D scene representation and a neural rendering pipeline.
It can rotate, translate, scale each object and change camera poses but with compromised image quality and resolution.

Compared to 2D generative models, these models take much more calculations to render an image (speed) and require much more memory during training to cache intermediate results (memory).
Computational constraints limit the rendering resolutions and quality.
For high-quality image synthesis (towards 512$\times$512 and beyond), recent methods turn to find more efficient and expressive representations of geometry and appearance, improve the training efficiency, or the combination of both. 
\revise{We focus on the strategies of learning efficient and expressive representations as follows in this subsection, leaving later parts to the next subsection.}

CIPS-3D~\cite{zhou2021cips3d} (Oct 2021) adopts a shallow NeRF network (containing only three SIREN blocks) to represent 3D shape and a deep 2D INR network to synthesis high-fidelity appearance. 
Inspired by recent progress in 3D surface reconstruction, GOF~\cite{xu2021generative} (Nov 2021)  combines implicit surfaces and radiance fields.
They reinterpret the alpha values 
in the rendering equation as occupancy representations and reformulate generative radiance fields by predicting alpha values instead of volume densities. %
\revise{EG3D~\cite{chan2022efficient} (Dec 2021) proposes to use a tri-plane hybrid 3D representation, formulated from explicit features of StyleGAN2 generator.}
They align explicit features along three axis-aligned orthogonal feature planes and query any 3D position by projecting it onto each of the three feature planes, resulting in an aggregated 3D feature.
The aggregated features are then interpreted as color and density by an additional lightweight decoding network. 
This tri-plane representation is 3 to 8 times faster than an implicit Mip-NeRF~\cite{barron2021mipnerf} network and only requires a fraction of its memory. A super-resolution module upsamples and refines raw neurally rendered images.
VolumeGAN~\cite{xu2022volumegan} (Dec 2021) explicitly learns a structural representation and a textural representation. 
It learns a feature volume to represent the underlying structure, which is transformed into a feature field based on a NeRF-style model. The feature field is then aggregated into a 2D feature map as the textural representation. A neural renderer is finally used for appearance synthesis.
StyleSDF~\cite{orel2022stylesdf} (Dec 2021) merges a SDF-based 3D representation into the 2D StyleGAN generator.
This framework consists of two main components: a backbone conditional SDF volume renderer and a StyleGAN generator. The renderer takes in a latent code and camera parameters, queries points and view directions within the volume, and projects 3D surface features onto 2D views. 
To overcome the drawbacks of GIRAFFE and inherit its 3D controllability, a follow-up work GIRAFFE-HD~\cite{xue2022giraffehd} (Mar 2022) leverages a style-based neural renderer, generates the foreground and background independently, and stitches them together to composite a coherent final image.  It enforces semantic disentanglement and 3D consistency through training constraints. 
In contrast to previou methods, %
VoxGRAF~\cite{schwarz2022voxgraf} (Jun 2022) adopts a 3D-aware GAN based on a sparse scene representation that allows for efficient rendering.
It parameterizes the radiance field on a sparse voxel grid rather than using a coordinate-based MLP and predicts colors and density values on this sparse voxel grid using volume rendering.

\subsubsection{Efficient and Consistent Rendering Algorithms}
\label{subsubsec:accelerated_and_consistent_rendering}

Early studies~\cite{nguyen2020blockgan, nguyen2019hologan} use simple 3D-to-2D projections for rendering. 
Such operations fail to produce high-quality images with fine-grained details and are restricted to representing poses in the training dataset.
For higher rendering quality, recent methods adopt the state-of-the-art neural volume rendering techniques.
GRAF~\cite{schwarz2020graf} and pi-GAN~\cite{chan2021pigan} implement a discretized form of the volume rendering equation and uses the stratified and hierarchical sampling approach introduced by NeRF. 
The neural volume rendering approach has several advantages over previous 3D-to-2D projections: 1) producing images with fine details and high resolutions and 2) allowing for explicit control over camera pose, focal length, aspect ratio, and other parameters.
Despite the advantages, the volume integrations approximated by sampling points along viewing rays are still costly for both training and inference.
Some methods render the 3D representation directly at the final image resolution~\cite{chan2021pigan,schwarz2022voxgraf}.
Due to the high memory and computation cost of volume rendering, direct rendering at target resolution is not efficient and struggles to generate images at high-resolution (512$\times$512 and beyond). 
Some recently-developed methods make use of a two-stage rendering process~\cite{niemeyer2021giraffe,gu2022stylenerf,orel2022stylesdf,chan2022efficient,xu2022volumegan} or develop efficient volume rendering strategy~\cite{pan2021shadegan,zhou2021cips3d,deng2022gram} for high-resolution image generation.
Meanwhile, aimed to reduce view-inconsistent artifacts brought by the 2D renderers, they adopt different strategies such as NeRF path regularization~\cite{gu2022stylenerf} and dual discriminators~\cite{chan2022efficient}.

\vspace{1mm} 
\noindent\textbf{Two-stage Rendering Process:}
To high-resolution image generation, some methods adopt a two-stage rendering process. 
Typically, they first generate a feature map at a low resolution and then employ upsampling in 2D space to progressively increase into the required high resolution. 
Niemeyer~\etal~\cite{niemeyer2021giraffe} improve training and rendering efficiency by combining NeRF with a ConvNet-based renderer.
Simlilarly, Chan~\etal~\cite{chan2022efficient} perform the majority of the training at a rendering resolution of 64$\times$64 and gradually increase the resolution, pixel-by-pixel, to 128$\times$128, which are fed into a super-resolution module to produce images at the target resolution.
However, these pixel-wise learnable upsamplers sacrifice view consistency and impair the quality of the learned 3D geometry due to network designs.
In contrast, non-learnable upsamplers that interpolate the feature map with pre-defined lowpass filters (\eg~bilinear interpolation) produce smoother results but lead to non-removable bubble artifacts.
In StyleNeRF~\cite{gu2022stylenerf}, the two approaches are combined to balance quality and consistency. 
Despite that upsampler scales the intermediate result to high resolution, it comes with two severe limitations: 1) the texture and shape change as camera moves; 2) the geometry is represented in a low resolution ($\approx 64^3$), both resulting in a compromised multi-view consistency of a generated object.
To overcome the above limitations of the two-stage rendering, Skorokhodov~\etal~\cite{skorokhodov2022epigraf} drop the upsampler and improve the patch-wise optimization strategy~\cite{schwarz2020graf} to build a 3D generator.
They redesign the discriminator by making it better suited to operating on image patches of variable scales and locations, along with changing the random scale sampling strategy from an annealed uniform to an annealed beta distribution.
This allows the proposed EpiGRAF to converge 2-3 times faster than upsampler-based architectures despite the generator modeling geometry in full resolution.

\vspace{1mm} 
\noindent\textbf{Efficient Rendering Strategy:} 
This part is categorized into two groups of approaches according to their adopted strategies. 
As for the first group, an expressive and efficient representation allows usage of a simplified sampling strategy~\cite{orel2022stylesdf} or a lightweight decoder~\cite{chan2022efficient}.
In StyleSDF~\cite{orel2022stylesdf}, using SDFs leads to higher view-consistency and expressiveness, even with a simplified volume sampling strategy. 
They sample $\mathrm{N}$ points from $\mathrm{N}$ evenly-sized bins of integration intervals instead of stratified sampling~\cite{schwarz2020graf,chan2021pigan,gu2022stylenerf,mildenhall2020nerf}, which reduces the number of samples by half.
VoxGARF~\cite{schwarz2022voxgraf} accelerates the rendering from the perspective of spatial sparsity, where volume rendering yields a foreground image and an alpha mask.
The second group aims to develop efficient sampling strategy~\cite{xu2021generative,pan2021shadegan,deng2022gram}. They constrain the point sampling in a reduced space rather than anywhere in the volume.
GRAM~\cite{deng2022gram} proposes a manifold predictor $\mathcal{M}$ to predict a reduced space for point sampling and radiance field learning.
$\mathcal{M}$ is a light-weight MLP that maps a point $\boldsymbol{x}$ to a scalar value $s$.
The predicted scalar field gives $N$ isosurfaces with a set of predefined levels (constant values $\{l_i\}$): 
$\mathcal{S}_i{=}\left\{\boldsymbol{x} {\mid} \mathcal{M}(\boldsymbol{x}){=}l_i\right\}$.
The rendering only samples points from intersections between the determined isosurfaces and a camera ray.
Pan~\etal~\cite{pan2021shadegan} introduce a light-weighted surface tracking network $S$ to estimate the rendered object surface. %
This saves rendering computations by just querying points near the predicted surface.
The sample region shrinks from the entire volume to a narrow interval around the surface.

\vspace{1mm} 
\noindent\textbf{Consistent View Regularization:}
Optimizing the radiance fields from a set of 2D training images can encounter critical degenerate solutions in the absence of geometry constraints, leading to a multi-view inconsistency problem in NeRF-based generative models.
Some methods propose multi-view regularizations on colors and shapes to improve photometric and geometric consistencies. 
Different strategies are adopted to reduce view-inconsistent artifacts brought by the 2D renderers~\cite{gu2022stylenerf,chan2022efficient,pan2021shadegan,shi2022improving,zhang2022multi}.
For example, Gu~\etal~\cite{gu2022stylenerf} propose a NeRF path regularization to enforce 3D consistency, which is implemented by sub-sampling high-resolution outputs and comparing them against the low-resolution image generated by NeRF.
Chan~\etal~\cite{chan2022efficient} propose a pose-conditioned dual discriminators with two modifications in traditional GAN discriminators. 
First, they pass the rendering camera intrinsics and extrinsics matrices to the discriminator as a conditioning label.
Second, they take as the input the concatenation of the final result $I_r$ and a low-resolution RGB image $I_r^{\prime}$. They interpret the first three feature channels of a neurally rendered feature image as $I_r^{\prime}$ and bilinearly upsample it to the same size of $I_r$.
This pose-conditioned dual discriminator is used in many follow-up studies.
Shi~\etal~\cite{shi2022improving} design a geometry-aware discriminator, GeoD, to improve 3D-aware GANs.
Besides the real/fake classification, they assign the discriminator an additional  geometry branch, aiming to derive the shape-related information (\eg~depth and normal), which is employed as an extra signal to supervise the generator.
Pan~\etal~\cite{pan2021shadegan} observe that small variations of shape could lead to similar RGB images that look equally plausible to the discriminator, as the color of many objects is locally smooth. 
This phenomenon is referred to as shape-radiance (color) ambiguity.
To eliminate this problem, they propose a multi-lighting constraint, which is realized by modeling illumination explicitly and rendering with various lighting conditions. 
To overcome the same issue, MVCGAN~\cite{zhang2022multi} (Apr 2022) builds geometry constraints by optimizing multiple views jointly to ensure geometry consistency between views. %
They minimize re-projection loss between a primary image and a warped image, and integrate a stereo mixup module to encourage the warped image to be similar to a real image. 
Such scheme guarantees geometry constraints between different views and supports large pose variations.
\revise{PoF3D~\cite{shi2023learning} (Jan 2023) notice a high sensitivity towards pose priors in existing 3D-aware image synthesis methods and propose a novel approach notably eliminating the need for pose priors.
It comprises two principal components: a pose-free generator and a pose-aware discriminator.
The pose-free generator maps a latent code to a neural radiance field as well as a camera pose. The pose-aware discriminator, on the other hand, first predicts a camera pose from the given image and then uses it as the pseudo label for conditional real/fake discrimination.
PoF3D demonstrates the potential for high-quality 3D-aware image synthesis without the reliance on 3D pose priors.}

\subsubsection{\revise{Broadening Applicable Scenarios}}
\label{subsubsec:broadening_applicable_scenarios}

\revise{Previous methods mainly focused on common categories, such as faces, cars, and other single categories with simple geometry and appearance. 
These methods can only generate a single canonical object and show limited capacity in composing a complex scene containing a variety of objects. 
However, recent methods have attempted to expand to more complex categories, such as the fine-grained shape~\cite{skorokhodov2022epigraf}, human bodies~\cite{zhang20213d,yang20223dhumangan}, and street scenes~\cite{xu2023discoscene}.
For example, EpiGRAF~\cite{skorokhodov2022epigraf} (Jun 2022) drops the upsampler and improves the patch-wise optimization strategy~\cite{schwarz2020graf} to build a 3D generator. The generator models the geometry in a full dataset resolution and is able to fit data where the global structure differs a lot between different objects in Megascans~\cite{skorokhodov2022epigraf}.
3D-SGAN~\cite{zhang20213d} (Nov 2022) and 3DHumanGAN~\cite{yang20223dhumangan} (Dec 2022) are two examples of 3D-aware human body synthesis. The human body exhibits a more diverse variety of shapes, poses, and texture variations, presenting far more significant challenges compared with modeling the human face.
DiscoScene~\cite{xu2023discoscene} (Dec 2022) is a 3D-aware generative model for scene synthesis.
The proposed model spatially disentangles the entire scene into object-centric generative radiance fields, leveraging only 2D images with global-local discrimination. DiscoScene not only achieves generation fidelity and editing flexibility for individual objects, but also efficiently composes these objects and the background into a complete scene.
There are other categories of 3D-aware generative models that require the inductive bias of specialized domain knowledge. For instance, significant progress has been made in the incorporation of 3D-aware image generation with the articulation of the body or face for controllable generative models. This approach combines 3D-aware generative models with deformation fields~\cite{hong2022eva3d,bergman2022generative,sun2023next3d}.
}

\revise{Notably, recent research efforts~\cite{skorokhodov20233d,sargent2023vq3d} have been made to extend 3D-aware image synthesis beyond a single category, aiming to encompass a wide range of categories, such as those found in ImageNet~\cite{deng2009imagenet}. 
Particularly, Skorokhodov~\etal~\cite{skorokhodov20233d} (Mar 2023) propose 3D generator with Generic Priors (3DGP). Building upon EpiGRAF~\cite{skorokhodov2022epigraf}, the generator first produces a tri-plane representation for the scene, given a random latent code $z$. Subsequently, a shallow 2-layer MLP predicts RGB color and density values from an interpolated feature vector at a 3D coordinate. Images and depths are volumetrically rendered at any given camera position.
The model is trained across all 1,000 classes of ImageNet~\cite{deng2009imagenet}, demonstrating the feasibility of multi-categorical 3D synthesis for non-alignable data.
}

\subsubsection{User-interactive Editing}
\label{subsubsec:interactive_editing}

Plenty of approaches are proposed to facilitate user-interactive editing.  They can change the background’s appearance independent of the foreground, translate or rotate the foreground object in 3D, and change the foreground object’s shape and color. 
For example, some methods~\cite{nguyen2020blockgan,niemeyer2021giraffe,xue2022giraffehd} use multiple noise vectors $z_i$ to represent the background and each foreground object. 
Unlike previous studies that learn 3D features directly for the whole scene~\cite{nguyen2019hologan}, 
they learn a 3D feature for each object, render separately, and stitch them together to composite a coherent final image.
Such design disentangles a scene into separate objects and enables control over camera pose, lighting, and shadow.

Differently, Kwak~\etal~\cite{kwak2022injecting} (Jul 2022) design a layer-wise SUbspace in INR NeRF-based generator (SURF-GAN).
Instead of being used represent the background and each foreground object, multiple noise vectors are injected layer-by-layer into NeRF-based SURF blocks, in which interpretable dimensions are captured in layers with sub-modulation vectors.
SURF-GAN~\cite{kwak2022injecting} includes several SURF blocks that take position and view direction as inputs to predict view dependent color.
IDE-3D~\cite{sun2022ide3d} (May 2022) enables local control of the facial shape and texture and supports real-time, interactive editing.
It makes two key modifications based on~\cite{chan2022efficient} to enable such interactive disentangled editing.
First, they take two (instead of one) codes respectively representing shape and texture, which gives 3D volumes of semantic and texture in the tri-plane representation. Second, they jointly render rgb images and semantic masks (instead of just rgb images) through the volume rendering. The dual discriminator is also changed accordingly to take as input the concatenation of rgb and semantic masks.

\revise{Recently, 3D GAN inversion methods have been developed~\cite{xie2023high,lan2023self,ko20233d}, which are based on the previously mentioned 3D GANs and are used for 3D-aware image editing.
There are two key distinctions between these 3D GAN inversion methods and 2D GAN inversion methods~\cite{zhu2020domain,xia2022gan}. Firstly, these 3D methods rely on a 3D-aware GAN model (\eg~EG3D~\cite{chan2022efficient}), instead of a 2D GAN (\eg~StyleGANs~\cite{karras2019style,karras2020analyzing}). Secondly, 3D GAN inversion methods take into account the camera pose.
The 3D GAN inversion methods not only achieve realistic and accurate manipulation but also excel in preserving the identity and geometry of the original input. These methods can generate edited 3D shapes from the modified latent codes, leveraging the capabilities of 3D-aware GANs. Additionally, these 3D GAN inversion methods deliver results that are comparable to those of other methods but necessitate significantly fewer computing resources when compared with training a 3D-aware image editing method.
}

\subsection{\revise{Unconditional 3D-aware Diffusion Models}}
\label{subsec:unconditional_3d_diffusion}

\revise{Recent trends in research also show a burgeoning interest in employing diffusion models~\cite{ho2020denoising,song2020denoising} for 3D generative modeling~\cite{zhang20233dshape2vecset,kim2023neuralfield,muller2023diffrf,anciukevivcius2023renderdiffusion,karnewar2023holodiffusion,xiang20233d,chan2023generative}.
Diffusion models, which have proven extremely effective in generating high-quality images in recent years, have not yet been widely applied to 3D-ware image synthesis. 
DiffRF~\cite{muller2023diffrf} (Dec 2022) introduces a novel method for 3D radiance fields synthesis, utilizing denoising diffusion probabilistic models. It learns multi-view consistent priors from posed image collections, enabling free-view image synthesis and precise shape generation. 
RenderDiffusion~\cite{anciukevivcius2023renderdiffusion} (Nov 2022) introduces a novel image denoising architecture that generates and renders an intermediate three-dimensional representation of a scene in each denoising step. This imposes a robust inductive structure within the diffusion process, yielding a 3D-consistent representation while only requiring 2D supervision. The resulting 3D representation can be rendered from any perspective.
HoloDiffusion~\cite{karnewar2023holodiffusion} (Mar 2023) introduces a novel diffusion framework that can be trained end-to-end with only posed 2D images for supervision. Furthermore, it proposes an image formation model that decouples model memory from spatial memory. 
3D-aware generative diffusion models are expected to catch up in the near future.
}

\subsection{Conditional 3D Generative Models}
\label{subsec:conditional_3d_generative_models}

\revise{The unconditional 3D generative models, primarily GANs (\cref{subsec:unconditional_3d_gans}) as well as some methods based on diffusion models (\cref{subsec:unconditional_3d_diffusion}), generally lack the capability to perform precise attribute editing for real images.
Similar to 2D counterparts, 3D-aware real image editing could either use pretrained 3D GANs via GAN inversion or train a 3D-aware generative model from scratch with additional inputs.
}

As an emerging technique to bridge the real and fake image domains, GAN inversion~\cite{zhu2020domain,xia2022gan} plays an essential role in enabling pretrained GAN models real image editing. 
It inverts a real image into the latent space of a trained GAN model, which allows us to alter image attributes by varying the inverted code in the latent space (known as latent space traversals).
For 3D-aware image editing, some rely directly on 2D GAN inversion and latent space traversal techniques, while others develop tools particularly for 3D GAN inversion.
For example, ShadeGAN~\cite{pan2021shadegan} (Oct 2021) could also be used to reconstruct a given image by performing GAN inversion. Such inversion with this method allows to obtain object properties from the image, such as shape, normal, albedo, and shading. 
FENeRF~\cite{sun2021fenerf} (Nov 2021) has attempted to edit the local shape and texture in a facial volume by using an optimization-based GAN inversion. 
Lin~\etal~\cite{lin20223d} (Mar 2022) propose a method for multi-view consistent video editing and animation based on 3D GAN inversion. They invert the video frames into the latent space of a pi-GAN by using pivotal tuning inversion (PTI)~\cite{roich2021pivotal} and edit face attributes by using StyleFlow~\cite{abdal2020styleflow}.
IDE-3D~\cite{sun2022ide3d} adopts a hybrid GAN inversion approach. 
Given a facial image and its semantic label, it obtains texture and semantic latent codes with corresponding encoders and uses them as the initialization for PTI to obtain high-fidelity reconstruction.
Editing is performed by drawing on inverted semantic masks.

Other methods target at pose-dependent views by training a 3D network conditioned on a single image. 
Pix2NeRF~\cite{cai2022pix2nerf} (Feb 2022) demonstrates that merely applying learning-based GAN inversion (learning an encoder and keeping the generator fixed during training) is insufficient to obtain an accurate mapping from image to latent space with pi-GAN as the backbone.
Instead, they train an encoder jointly with a generator and a discriminator (both of the same architecture and procedure as in~\cite{chan2021pigan}). Once trained, given an input image, Pix2NeRF disentangles its pose and content and renders novel views of the content.
Sem2NeRF~\cite{chen2022sem2nerf} (Mar 2022) takes as input a single-view 2D semantic mask and outputs a NeRF-based 3D representation that can be used to render photorealistic images in a 3D-aware view-consistent manner.
AutoRF~\cite{muller2022autorf} (Apr 2022) focuses on novel view synthesis of objects without background. This model consists of an encoder that extracts a shape and an appearance code from an object’s image, which can be decoded into an implicit radiance field operating in normalized object space and leveraged for novel view synthesis. Object images are generated from real-world imagery by leveraging machine-generated 3D object detections and panoptic segmentation. 

\section{Discussion}
\label{sec:future_direction}

\revise{Despite great advances in deep 3D-aware generative image synthesis, challenges remain and its rapid growth is expected to continue. In the following, we provide an overview of future directions, problems to solve, and trends to anticipate. Due to their unique characteristics, some limitations or future trends may only apply to certain categories of methods.}

\vspace{1mm} 
\textbf{Quality:} 
Unlike traditional graphic rendering, since implicit neural representations do not provide an explicit and holistic 3D shape for rendering, inconsistencies seems inevitable in the generated surface and texture under different viewpoints. 
Moreover, these strategies are introduced separately rather than endogenously as part of the method.
We expect that in the future there will be more endogenous approaches to 3D-aware generative models that generates high-quality, high-resolution, multi-view-consistent images in real time, as well as high-quality 3D geometry.

\vspace{1mm} 
\textbf{Speed:} 
There is typically a slow training and inference speed with 3D-aware image synthesis methods. 
Most attempts to accelerate training and inference time require extra memory for caching trained models or use additional voxel/spatial-tree based scene features. It is expected that future speed-based methods should develop memory-friendly frameworks as well as novel inclusive and learnable scene representations to accelerate training and inference.

\vspace{1mm} 
\textbf{Editability:} 
In spite of the promising quality of the produced results, most methods are incapable of editing individual image components. In some methods, latent vectors are used at various points along the pipeline to control composition, shape, and appearance of images. The latent codes enable the model to control small changes in scene content, such as lighting or coloration, per image. Others allow additional input to change aspects of the scene, such as images, texts, semantic labels, or direct control parameters.
\revise{The editability of deep 3D-aware generative image synthesis methods, however, still has plenty of room for improvement compared with their 2D counterparts.}

\vspace{1mm} 
\textbf{Forensics:} 
The success of recent generative models has led to many new applications, but also raised ethical and social concerns, such as fraud and fabricated images, videos, and news (known as deepfakes). The ability to detect deepfakes is essential to preventing malicious usage of these models. Recent studies have shown that a classifier can be trained to distinguish deepfakes and generalize to unseen architectures. It may continue to be a cat-and-mouse game in the future, since generated images will become increasingly difficult to detect. Conversely, these images can also be utilized as the training data for identifying fakes.

\vspace{1mm} 
\revise{\textbf{Generalization:} 
3D-aware generative image synthesis has been limited to a narrow range of perspectives, falling short of the free-view rendering capabilities seen in NVS. Recent efforts, inspired by progress in INR-based NVS, have expanded the range of camera movements, but still lack the ability to generate 360-degree views. Additionally, while there have been attempts to broaden 3D-aware image synthesis from common categories, which are characterized by simple geometry and appearance, to intricate scenes composed of various objects, these attempts often fall short in terms of controllability and quality. 
Therefore, there exists a substantial opportunity for future research to broaden the scope of 3D-aware image synthesis to cover a more diverse range of categories applicable in real-world scenarios. However, this expansion will inevitably increase computational complexity, which must be adequately addressed.
}

\vspace{1mm} 
\textbf{Network Design:}
Most current methods rely on a 2D architecture design that introduces 3D representation, rendering, and multi-view regularization to make a 2D model 3D-aware. The majority of 3D-aware generative models use convolution-powered generative adversarial networks. In recent years, transformer-based 2D image synthesis methods
have emerged corresponding to their counterparts in convolutional networks.
Developing 3D-aware generative models that are built on transformers with lighter structures and lower computational demands remains to be explored. Since almost all the methods above are based on GANs, developing other generative models that are 3D-aware, especially diffusion models, is also a promising future direction.%

\vspace{1mm} 
\textbf{Evaluation metrics:}
It remains to be explored whether there are any reliable metrics that can better evaluate the photorealistic and geometric quality of generated images. 
Image quality and diversity are mainly measured by general metrics used for generative models. 
\revise{3D consistency is often evaluated by measuring distances between (pseudo-)depth maps of generated images or comparing similarities of generated face identities at different camera positions. Some use COLMAP reconstruction on their rendering outputs to demonstrate the 3D consistency.}
However, considering the lack of real reference, these measures only partially reflect the stability of generated results, but cannot reflect the distance from real samples.
There is still a lack of effective assessment tools to evaluate the difference between the predicted and expected outcomes in a more reliable and direct manner for deep generative 3D-aware image synthesis.

\section{Conclusion}
\label{sec:conclusion}
\revise{This paper presents a comprehensive overview of recent advances in deep 3D-aware generative image synthesis. 
We propose a systematic taxonomy for deep 3D-aware generative image synthesis methods. Specifically, we categorize the existing approaches into two groups: 3D control of 2D generative models and 3D-aware generative models.
We also identify some open problems on this topic to inspire future research.
We hope that this timely and up-to-date survey will serve as a starting point for future research to help advance this emerging and challenging field.
}

\begin{acks}
This work was supported by the Engineering and Physical Sciences Research Council [grant number EP/W523835/1].
\end{acks}

\bibliographystyle{ACM-Reference-Format}
\bibliography{reference}

\end{document}

%% file: command/figure.tex
\newcommand{\figoverview}{
\begin{figure*}[t!]
\renewcommand{\arraystretch}{1.3}
\centering
\footnotesize
\includegraphics[width=0.85\linewidth]{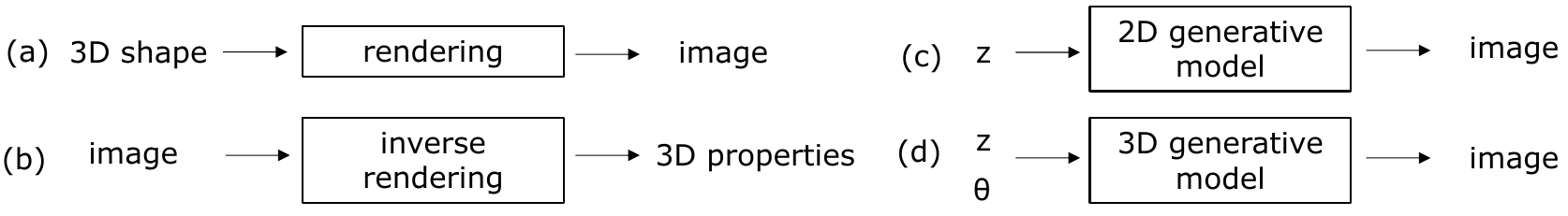}
\caption{Comparison of (a) rendering, (b) inverse rendering, (c) 2D generative models, and (d) 3D generative models. 3D generative models learn 3D representations first and then render a 2D image at certain viewpoints. Both 2D and 3D generative models have unconditional and conditional settings. An unconditional generative model maps a noise input $z$ (and a camera pose in 3D models) to a fake image;
a conditional model takes additional inputs as control signals, which could be image, text, or a categorical label.}
\label{fig:overview}
\vspacefigtext
\end{figure*}
}
\newcommand{\figtaskillustration}{
\begin{figure*}[t!]
\renewcommand{\arraystretch}{1.3}
\centering
\footnotesize
\includegraphics[width=\linewidth]{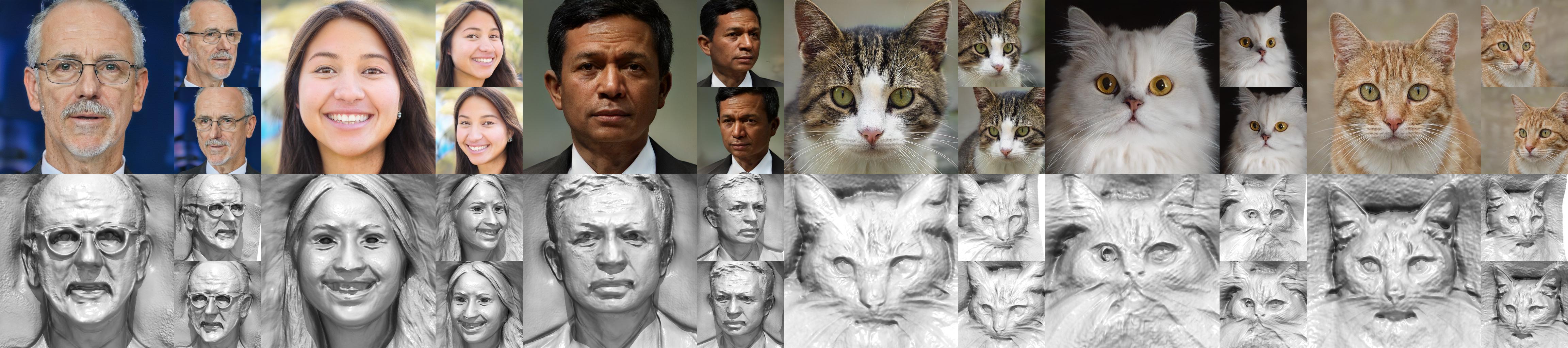}
\caption{\revise{The illustrative examples of 3D-aware image synthesis demonstrate the objective of this task, which is to generate high-quality renderings that are consistent across multiple views (top) and potentially provide detailed geometry (bottom). Typically, deep generative 3D-aware image synthesis methods are trained using a collection of 2D images, without depending on target-specific shape priors, ground truth 3D scans, or multi-view supervision. Examples are sourced from~\cite{chan2022efficient}.}
}
\label{fig:taskillustration}
\vspacefigtext
\end{figure*}
}

\newcommand{\tabsingledataset}{
\setlength{\tabcolsep}{2pt}
\begin{table*}[t]
    \centering
    \caption{Summary of popular single-view image datasets organized by their major categories and roughly sorted by their popularity.}
    \label{tab:singledata}
    \resizebox{0.85\linewidth}{!}{
    \scriptsize
    \begin{tabular}{l l l c c c l} 
        \toprule
        \begin{tabular}{@{}c@{}}\textbf{dataset} \\  \end{tabular} 
        & \begin{tabular}{@{}c@{}} \textbf{published in} \\  \end{tabular} 
        & \begin{tabular}{@{}c@{}} \textbf{category} \\  \end{tabular}
        & \begin{tabular}{@{}c@{}} \textbf{\# samples} \\  \end{tabular}
        & \begin{tabular}{@{}c@{}} \textbf{resolution} \\  \end{tabular}
        & \begin{tabular}{@{}c@{}} \textbf{representative references} \\  \end{tabular}
        & \begin{tabular}{@{}c@{}} \textbf{keyword} \\  \end{tabular}
        \\ 
        \midrule
        FFHQ~\cite{karras2019style} & CVPR 2019 & Human Face & 70k & 1024 $\times$ 1024 &\cite{orel2022stylesdf,xu2022volumegan,gu2022stylenerf,chan2022efficient} & single, simple-shape\\ 
        AFHQ~\cite{choi2020starganv2} & CVPR 2020 & Cat, Dog, and Wildlife & 15k & 512 $\times$ 512 &\cite{orel2022stylesdf,xu2022volumegan,gu2022stylenerf,chan2022efficient} & single, simple-shape\\
        \revise{SHHQ~\cite{fu2022stylegan}} & ECCV 2022 & Human Body & 230k & 1024 $\times$ 512 &\cite{yang20223dhumangan} & single, variable-shape\\
        CompCars~\cite{yang2015large} & CVPR 2015 & Real Car & 136K & 256$\times$256 &\cite{xu2022volumegan,gu2022stylenerf,nguyen2019hologan,nguyen2020blockgan} & single, simple-shape\\
        CARLA~\cite{dosovitskiy2017carla} & CoRL 2017 & Synthetic Car & 10k & 128$\times$128 &\cite{chan2021pigan,deng2022gram,xu2022volumegan,tewari2022disentangled3d,cai2022pix2nerf} & single, simple-shape\\
        CLEVR\textit{n}~\cite{johnson2017clevr} & CVPR 2017 & Objects & 100k & 256$\times$256 &~\cite{nguyen2020blockgan,niemeyer2021giraffe} & multiple, simple-shape\\ 
        LSUN~\cite{yu2015lsun} & 2015 & Bedroom & 300K & 256 $\times$ 256 &\cite{xu2022volumegan,nguyen2019hologan} & single, simple-shape\\
        CelebA~\cite{liu2015faceattributes} & ICCV 2015 & Human Face & 200k & 178$\times$218 &\cite{nguyen2019hologan,xu2022volumegan} & single simple-shape\\
        CelebA-HQ~\cite{karras2017progressive} & ICLR 2018 & Human Face & 30k & 1024 $\times$ 1024 &\cite{rebain2022lolnerf} & single, simple-shape\\
        MetFaces~\cite{karras2020ada} & NeurIPS 2020 & Art Face & 1336 & 1024 $\times$ 1024 &\cite{gu2022stylenerf} & single, simple-shape\\
        M-Plants~\cite{skorokhodov2022epigraf} & NeurIPS 2022 & Plant & 141,824 & 256 $\times$ 256 &\cite{skorokhodov2022epigraf} & single, variable-shape\\
        M-Food~\cite{skorokhodov2022epigraf} & NeurIPS 2022 & Food & 25,472 & 256 $\times$ 256 &\cite{skorokhodov2022epigraf} & single, variable-shape\\
    \bottomrule
    \end{tabular}
    }
    \vspacefigtext
\end{table*}
}

\newcommand{\tabquangans}{
\begin{table*}[thbp]
    \centering
    \footnotesize
    \SetTblrInner{rowsep=1.0pt}       %
    \SetTblrInner{colsep=1.8pt}       %
    \caption{\revise{The performance details of several representative unconditional 3D-aware GANs on the FFHQ dataset~\cite{karras2019style}, with the FID used as the evaluation metric.}}
    \label{tab:quantitative_conditional_3d_gans}
    \vspace{3pt}
    \begin{tblr}{
        cells={halign=c,valign=m},  %
        column{1}={halign=l},       %
        cell{4,7}{1,2,3,4}={r=3}{},
        cell{013}{1,2,3,4}={r=2}{},
        hline{1,2,3,4,7,10,11,12,13}={1-7}{},      %
        hline{1,15}={1.0pt},         %
        hline{2}={0.7pt},            %
        vline{2,3,4,5,6}={1-14}{},       %
    }
        \textbf{Method} & \textbf{Publication} & \textbf{Renderer} & \textbf{Upsampler} &  \textbf{Resolution} & \textbf{FID on FFHQ}   \\
        GRAF~\cite{schwarz2020graf}      &  NeurIPS 2020    & Density, Color           & No  & 128$\times$128   &  46.30 \\
        $\pi$-GAN~\cite{chan2021pigan}   & CVPR 2021  & Density, Color           & No  & 128$\times$128   &  29.90  \\
        StyleNeRF~\cite{gu2022stylenerf}  & ICLR 2022 & Density, Color & Yes & 256$\times$256   &  8.00       \\
                                   &&&                                                                    & 512$\times$512   &  7.80 \\
                                   &&&                                                                    & 1024$\times$1024 &  8.10 \\
        StyleSDF~\cite{orel2022stylesdf}  & CVPR 2022  & SDF, Color      & Yes & 256$\times$256   &  11.50 \\
                                   &&&                                                                     & 512$\times$512   &  10.07 \\
                                   &&&                                                                     & 1024$\times$1024 &  10.01 \\
        VolumeGAN~\cite{xu2022volumegan}  & CVPR 2022 & Density, Color & Yes & 256$\times$256   &  9.10 \\
        GRAM~\cite{deng2022gram}      & CVPR 2022 & Occupancy, Color	      & No  & 256$\times$256   &  14.50     & \\
        EpiGRAF~\cite{skorokhodov2022epigraf}      & NeurIPS 2022 & Density, Color           & No  & 512$\times$512   &  9.92   \\
        EG3D~\cite{chan2022efficient}            & CVPR 2022 & Density, Color   & Yes & 256$\times$256   &  4.80      \\
                                   &&&                                                                    & 512$\times$512   &  4.70 \\
    \end{tblr}
\end{table*}
}

\newcommand{\figcomparison}{
\begin{figure*}[t!]
\renewcommand{\arraystretch}{1.3}
\centering
\footnotesize
\includegraphics[width=0.95\linewidth]{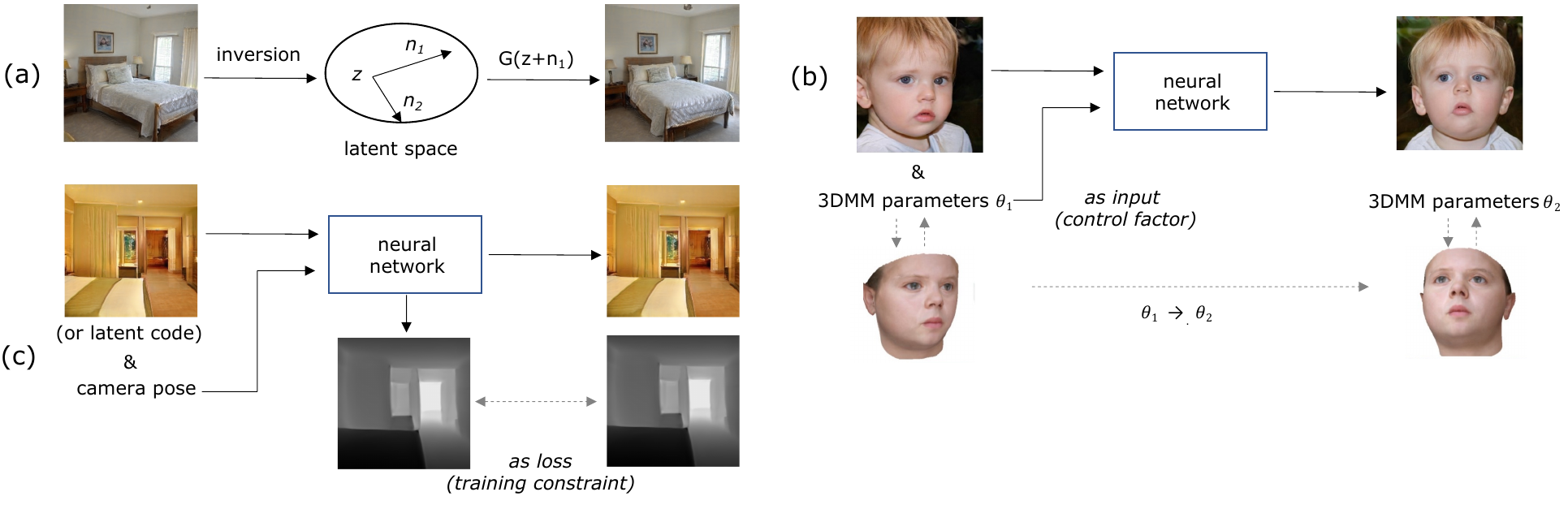}
\caption{Comparison of methods aiming to make 2D generative models (mostly GANs) 3D aware. These works are categorized into three groups based on how the 3D control capability is introduced: a) exploring 3D control in the 2D pretrained GAN latent spaces, b) using 3D parameters (\eg~from 3DMM~\cite{blanz1999morphable}) as input for explicit control, and c) introducing 3D-aware components (\eg~depth estimation module) into 2D GANs and adopting 3D prior knowledge (\eg~depth map) as training constraints.
}
\label{fig:group_1_comparison}
\vspacefigtext
\end{figure*}
}

\newcommand{\figdcomparison}{
\begin{figure*}[t!]
\renewcommand{\arraystretch}{1.3}
\centering
\footnotesize
\includegraphics[width=0.95\linewidth]{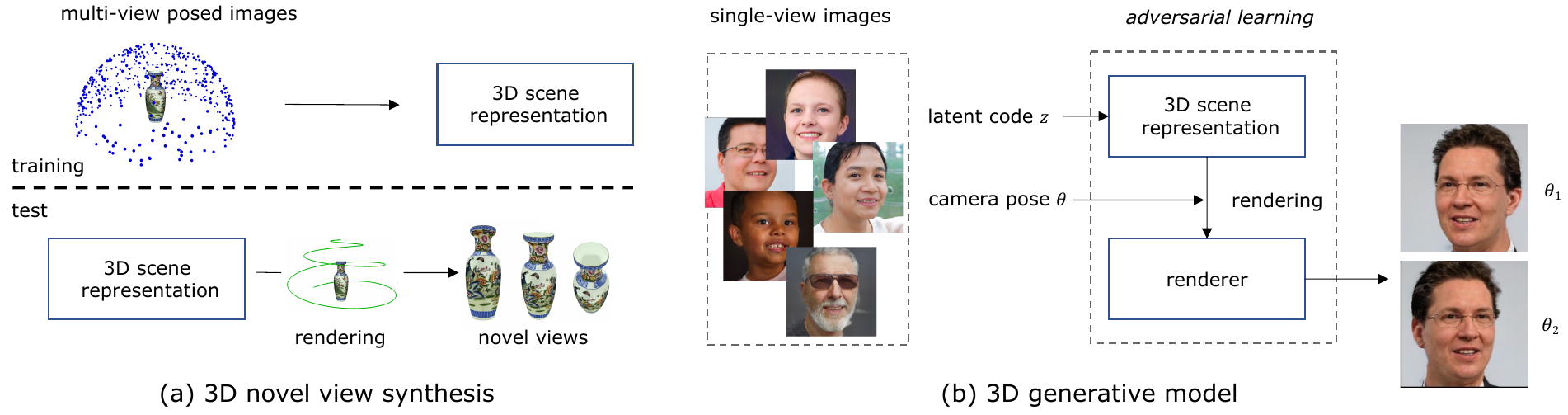}
\caption{\revise{The clarification of the defined terminology and the differences between 3D novel view synthesis methods (\cref{subsec:3d_novel_view_synthesis}) and 3D-aware generative models (\cref{sec:3d_generative_models}).
Both aim to generate multi-view-consistent and photo-realistic images, using a similar pipeline that learns the 3D representation first and then renders it.
Their differences lie in the application scenarios and training data. 3D novel view synthesis (a) learns a 3D representation from \textit{multiple views} of a scene. 3D generative models (b) learn to generate images from a collection of \textit{single views}.
Once trained, 3D-aware generative models generate images under a limited range of viewpoints while NVS methods allow for the free-view rendering of high-fidelity and detailed images. 
}
}
\label{fig:group_23_comparison}
\vspacefigtext
\end{figure*}
}

\newcommand{\tabexplicitcontrol}{
\setlength{\tabcolsep}{2pt}
\begin{table}[t!]
    \centering
    \caption{Overview of methods that incorporate 3D parameters to a 2D generative model.}
    \resizebox{0.75\linewidth}{!}
    {
    \scriptsize
    \begin{tabular}{l l l l} 
        \toprule
        \begin{tabular}{@{}c@{}}\textbf{Method} \\  \end{tabular}  
        & \begin{tabular}{@{}c@{}} \textbf{Publication} \\  \end{tabular} 
        & \begin{tabular}{@{}c@{}} \textbf{Control Factor} \\  \end{tabular} 
         & \begin{tabular}{@{}c@{}} \textbf{Supervision ($\theta \leftrightarrows I$)
         } \\  \end{tabular} 
        \\ 
        \midrule
        StyleRig~\cite{tewari2020stylerig} & CVPR 2020  & 3DMM parameters; $\theta$ & $I=
        \mathcal{R}(\theta)$ \\
        DiscoFaceGAN~\cite{deng2020disentangled} & CVPR 2020 & 3DMM, SH, angle vector; $\theta$ & $\theta=\mathcal{F}(I)$ \\
        PIE~\cite{tewari2020pie} &TOG 2020  & 3DMM parameters; $\theta$ & $\theta=\text{MoFa}(G(w))$ \\
        CONFIG~\cite{kowalski2020config} & ECCV 2020 & graphic parameters $\theta$ & synthetic images with $\theta$ \\
        GAN-Control~\cite{shoshan2021gan} & ICCV 2021 & \revise{intuitive representation} $y$ &
        $y_i = \mathcal{R}_i(G(z))$  \\
        3D-FM GAN~\cite{liu20223d} & ECCV 2022 & a rendering $I_r(\theta)$ (equiv. to $\theta$) & (image, rendering) pairs \\
        \bottomrule
    \end{tabular}
    }
    \label{tab:explicitcontrol}
    \vspacefigtext
\end{table}
}

\newcommand{\taballcomparison}{
\setlength{\tabcolsep}{2pt}
\begin{table*}[thbp]
    \centering
    \caption{\revise{Overview of deep generative 3D-aware image synthesis methods. 
    "Dataset" means what \textit{Categories} of images used for training.
    "Posed images" mean 2D images along with respective extrinsic and intrinsic camera matrices.
    "Geometry (Geo.)" indicates if the model is available for geometry reconstruction.
    "Controllability (Cont.)" means if the model has the ability of attribute edit beyond camera pose.
    "Condition" refers to additional inputs accompanying the latent code, while "Unconditional (Uncon.)" indicates the absence of any extra input. "Unsupervised (Unsup.)" implies that there is no additional training supervision apart from the images.}
    }
    \resizebox{\linewidth}{!}
    {
    \scriptsize
    \begin{tabular}{|l|l|c|c|c|c|l|} 
        \toprule
        \begin{tabular}{@{}c@{}}\textbf{Method} \\  \end{tabular}  
        & \begin{tabular}{@{}c@{}} \textbf{Publication} \\  \end{tabular} 
        & \begin{tabular}{@{}c@{}} \textbf{Dataset / Category} \\  \end{tabular} 
        & \begin{tabular}{@{}c@{}} \textbf{Condition} \\  \end{tabular} 
        & \begin{tabular}{@{}c@{}} \textbf{Geo.} \\  \end{tabular} 
        & \begin{tabular}{@{}c@{}} \textbf{Cont.} \\  \end{tabular} 
        & \begin{tabular}{@{}c@{}} \textbf{Supervision} \\  \end{tabular} 
        \\ 
        \aboverulesepcolor{gray!30!} \midrule \belowrulesepcolor{gray!30!}
        \rowcolor{gray!30!} 
        \multicolumn{7}{c}{
        \textbf{3D Control in 2D Latent Spaces (\cref{subsec:exploring_3d_control_in_2d_latent_space})}
        } 
        \\ \aboverulesepcolor{gray!30!}        
        \midrule
        GANSteerability~\cite{jahanian2020steerability} & ICLR 2020 & ImageNet & N/A & \nxmark & \ncmark & shifted image\\
        GANLatentDiscovery~\cite{voynov2020latent} & ICML 2020 & Face, ILSVRC & N/A & \nxmark & \ncmark & Unsup.\\
        InterFaceGAN~\cite{shen2020interpreting} & CVPR 2020 & Face & N/A & \nxmark & \ncmark & synthetic image \& label\\ 
        GANSpace~\cite{eric2020GANSpace} & NeurIPS 2020 & Face, ImageNet & N/A & \nxmark & \ncmark & Unsup.\\    
        SeFa~\cite{shen2021closedform} & CVPR 2021 & Face, Car, LSUN, ImageNet & N/A & \nxmark & \ncmark & closed-form\\ 
       \revise{LatentCLR}~\cite{yuksel2021latentclr} & ICCV 2021 & Face, Car, Cat, Bird & N/A & \nxmark & \ncmark & Unsup.\\ 
        \midrule \belowrulesepcolor{gray!30!}
        \rowcolor{gray!30!}\multicolumn{7}{c}{\textbf{3D Parameters as Controls (\cref{subsec:3d_parameter_as_controls})}} \\ \aboverulesepcolor{gray!30!} \midrule
        StyleRig~\cite{tewari2020stylerig} & CVPR 2020  & Face & synthesized image & \nxmark & \ncmark & 3DMM\\
        DiscoFaceGAN~\cite{deng2020disentangled} & CVPR 2020 & Face & parameters & \nxmark & \ncmark & 3DMM \\
        PIE~\cite{tewari2020pie} &TOG 2020 & Face  & real image & \nxmark & \ncmark & 3DMM  \\
        CONFIG~\cite{kowalski2020config} & ECCV 2020 & Face  & image \& parameters & \nxmark & \ncmark & synthetic data  \\
        GAN-Control~\cite{shoshan2021gan} & ICCV 2021 & Face & parameters & \nxmark & \ncmark & pseudo param. \\
        3D-FM GAN~\cite{liu20223d} & ECCV 2022 & Face & image \& rendering & \nxmark & \ncmark  & synthetic data  \\ 
        \midrule \belowrulesepcolor{gray!30!}
        \rowcolor{gray!30!}\multicolumn{7}{c}{\textbf{3D Priors as Constraints (\cref{subsec:3d_prior_as_constraints})}} \\ \aboverulesepcolor{gray!30!} \midrule
        $S^2$-GAN~\cite{wang2016generative} & ECCV 2016 & scene (NYUv2) & Uncon. & \nxmark & \nxmark & normal map \\ 
        PrGANs~\cite{gadelha20173d} &3DV 2017 & synthesized from 3D shapes & Uncon. & \ncmark & \nxmark & image and viewpoint  \\        
        VON~\cite{zhu2018visual} &NeurIPS 2018 & Chair, Car & Uncon. & \ncmark & \nxmark & 2.5D sketch, 3D shape  \\
        RGBD-GAN~\cite{noguchi2020rgbd} & ICLR 2020 & Face, Car & Uncon. & \ncmark  & \nxmark & Unsup.  \\ 
        NGP~\cite{chen2021towards} & CGF 2021 & Chair, Car & user controls & \ncmark & \ncmark & 3D shape, reflectance map  \\
        LiftedGAN~\cite{shi2021lifting} & CVPR 2021 & Face & Uncon. & \ncmark & \ncmark & pseudo multi-view images  \\
        DepthGAN~\cite{shi20223d} & ECCV 2022 & Scene (LSUN) & Uncon. & \ncmark & \ncmark & pseudo depth map \\   
        \midrule \belowrulesepcolor{gray!30!}
        \rowcolor{gray!30!}\multicolumn{7}{c}
        {
        \textbf{\revise{3D GANs (\cref{subsec:unconditional_3d_gans} and \cref{subsec:conditional_3d_generative_models})}}
        } \\ \aboverulesepcolor{gray!30!} \midrule
        HoloGAN~\cite{nguyen2019hologan} & ICCV 2019 & Face, Cat, Car, LSUN & Uncon. & \nxmark & \nxmark & Unsup. \\
        Liao~\etal~\cite{liao2020towards} & CVPR 2020 & Multiple Objects & Uncon. & \ncmark & \nxmark  & Unsup. \\
        BlockGAN~\cite{nguyen2020blockgan}  & NeurIPS 2020 & Multiple Objects & Uncon. & \nxmark & \ncmark & Unsup. \\
        GRAF~\cite{schwarz2020graf} & NeurIPS 2020 & rendered chair, Face, Cat, bird & Uncon. & \nxmark & \nxmark & Unsup. \\
        pi-GAN~\cite{chan2021pigan} & CVPR 2021 & Face, Car, CARLA & Uncon. & \ncmark & \nxmark & Unsup. \\
        GIRAFFE~\cite{niemeyer2021giraffe} & CVPR 2021 & Face, Cat, Car, Church, Chair & Uncon. &\nxmark & \ncmark & Unsup.  \\
        GOF~\cite{xu2021generative}  & NeurIPS 2021 & Face, Cat, Car & Uncon. & \ncmark & \nxmark & Unsup. \\
        ShadeGAN~\cite{pan2021shadegan} & NeurIPS 2021 & Face, Cat &  Uncon. & \ncmark & \nxmark & Unsup. \\
        CAMPARI~\cite{niemeyer2021campari} & 3DV 2021 & Face, Cat, Car, Chair & Uncon. & \nxmark & \ncmark & Unsup. \\
        StyleNeRF~\cite{gu2022stylenerf} & ICLR 2022 & Face, Cat, Car & Uncon. & \ncmark & \nxmark & Unsup. \\
        GRAM~\cite{deng2022gram} & CVPR 2022 & Face, Cat, CARLA & Uncon. & \nxmark & \nxmark & Unsup. \\
        EG3D~\cite{chan2022efficient} & CVPR 2022 & Face, Cat & Uncon. & \ncmark & \nxmark & Unsup. (posed 2D image) \\
        VolumeGAN~\cite{xu2022volumegan} & CVPR 2022 & Face, Cat, Car, bedroom, CARLA & Uncon. & \ncmark & \nxmark & Unsup. \\
        StyleSDF~\cite{orel2022stylesdf} & CVPR 2022 & Face, Cat & Uncon. & \ncmark & \nxmark  & Unsup. \\
        Pix2NeRF~\cite{cai2022pix2nerf} & CVPR 2022 & Face, CARLA, rendered image & image & \nxmark & \nxmark & Unsup. \\
        Sem2NeRF~\cite{chen2022sem2nerf} & ECCV 2022 & Face, Cat & semantic mask & \nxmark & \nxmark & semantic mask \& image\\
        SURF-GAN~\cite{kwak2022injecting} & ECCV 2022 & Face & Uncon. & \nxmark & \nxmark & Unsup. \\
        EpiGRAF~\cite{skorokhodov2022epigraf} & NeurIPS 2022 & Face, Cat, variable-shape & Uncon. & \ncmark & \nxmark & Unsup. \\
        IDE-3D~\cite{sun2022ide3d} & TOG 2022 & Face & Uncon. & \ncmark & \ncmark & semantic mask \& image \\
        \revise{3D-SGAN}~\cite{zhang20213d}  & ECCV 2022 & Human Body & Uncon. & \nxmark & \ncmark & semantic mask \& image \\
        \revise{DiscoScene}~\cite{xu2023discoscene} & CVPR 2023 & Street Scene & Uncon. & \nxmark & \nxmark & Unsup. \\
        \revise{3DGP}~\cite{skorokhodov20233d}  & ICLR 2023 & ImageNet, SDIP & Uncon. & \ncmark & \nxmark & Unsup. \\
        \midrule \belowrulesepcolor{gray!30!}
        \rowcolor{gray!30!}\multicolumn{7}{c}
        {\textbf{\revise{3D Diffusion Models (\cref{subsec:unconditional_3d_diffusion})}}
        } \\ \aboverulesepcolor{gray!30!} \midrule
        \revise{DiffRF~\cite{muller2023diffrf}} & CVPR 2023 & Chair & Uncon. /image & \ncmark & \ncmark &  Unsup. (posed 2D image) \\ \revise{RenderDiffusion~\cite{anciukevivcius2023renderdiffusion}} & CVPR 2023 & Face, CLEVR, ShapeNet & Uncon. /image & \nxmark & \nxmark & Unsup. (posed 2D image) \\  \revise{HoloDiffusion~\cite{karnewar2023holodiffusion}} & CVPR 2023 & object (CO3D) & Uncon. & \nxmark & \nxmark & Unsup. (posed 2D video) \\    
        \bottomrule
    \end{tabular}
 }   
    \label{tab:allcomparison}
    \vspacefigtext
\end{table*}
}

%% file: command/tree.tex
\begin{figure}[t!]
\centering
\tikzset{
    my node/.style={
        font=\small,
        rectangle,
        draw=#1!75,
        align=justify,
    }
}
\forestset{
    my tree style/.style={
        for tree={grow=east,
            parent anchor=east, %
            child anchor=west,  %
        where level=0{my node=black,text width=1.5em}{},
        where level=1{my node=black,text width=8.3em}{},
        where level=2{my node=black,text width=9.6em}{},
        where level=3{my node=black,text width=17.5em}{},
            l sep=1.5em,
            forked edge,                %
            fork sep=1em,               %
            edge={draw=black!50, thick},                
            if n children=3{for children={
                    if n=2{calign with current}{}}
            }{},
            tier/.option=level,
        }
    }
}
\scalebox{0.75}{
    \begin{forest}
      my tree style
[All
    [3D-aware Generative Models (\cref{sec:3d_generative_models})
        [Conditional 3D-aware Generative Models (\cref{subsec:conditional_3d_generative_models})
            [Textual Description]
            [Semantic Label]
            [Image]
        ]
        [Unconditional 3D-aware Diffusion Models (\cref{subsec:unconditional_3d_diffusion})
        ]
        [Unconditional 3D-aware GANs (\cref{subsec:unconditional_3d_gans})
            [User-interactive Editing (\cref{subsubsec:interactive_editing}), draw, dashed]
            [Broadening Applicable Scenarios (\cref{subsubsec:broadening_applicable_scenarios}), draw, dashed]
            [Efficient and Consistent Rendering (\cref{subsubsec:accelerated_and_consistent_rendering})
            ]
            [Efficient and Effective Representations (\cref{subsubsec:efficient_and_expressive_3d_representation})]
        ]
    ] 
    [3D Control of 2D Generative Models (\cref{sec:3d_control_of_2d_generative_models})
        [3D Priors as Constraints (\cref{subsec:3d_prior_as_constraints})
            [3D Components into 2D GANs (\cref{subsubsec:introducing_3d_components_into_2d_gans})]
            [3D Prior Knowledge (\cref{subsubsec:3d_prior_knowledge}), draw, dashed]
        ]
        [3D Parameters as Controls (\cref{subsec:3d_parameter_as_controls})
            [Explicit Control over 3D Parameters (\cref{subsubsec:explicit_control_over_3d_parameters})]
            [Factors from Pretrained Models (\cref{subsubsec:control_factors}), draw, dashed]
        ]
        [3D Control in 2D Latent Spaces (\cref{subsec:exploring_3d_control_in_2d_latent_space})
            [Pinpointing Predetermined Targets (\cref{subsubsec:pinpointing_predetermined_destination})]
            [3D Control Latent Directions (\cref{subsubsec:discovering_3D_control_latent_directions})]
        ]
    ] 
]
\end{forest}
}
\centering
\caption{\comments{A systematic taxonomy proposed in this survey of deep generative 3D-aware image synthesis methods.
The dashed borders at the third level denote preliminaries, applications, or issues discussed in this subcategory.
It should be noted that these methods are not mutually exclusive. For example, a few methods introduce 3D parameters to improve controllability (\cref{subsec:3d_parameter_as_controls}) while also implementing 3D constraints to improve consistency across multiple views (\cref{subsec:3d_prior_as_constraints}); EG3D~\cite{chan2022efficient} is referenced in \cref{subsubsec:efficient_and_expressive_3d_representation} and \cref{subsubsec:accelerated_and_consistent_rendering} for its approach to 3D-aware representations and rendering algorithms.          
}}
\label{fig:phylogenetic_tree}
\end{figure}
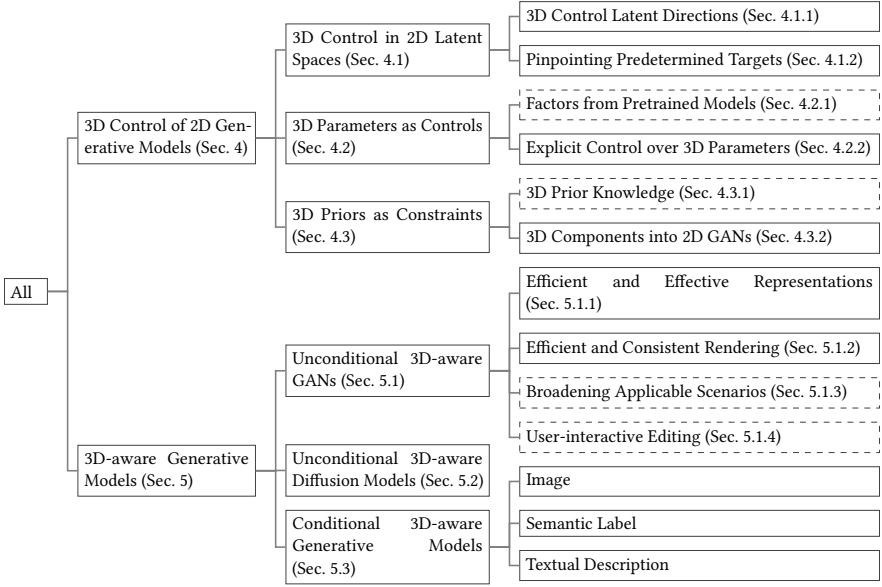

%% file: command/timeline.tex
\begin{figure*}[t!]
\centering
\resizebox{0.90\textwidth}{!}{%
\begin{tikzpicture}%
  \newcount\yearOne; 
  \yearOne= 2019%
  \def\n{4} %
  \def\w{18}       %
  \def\lt{0.40}    %
  \def\lf{0.36}    %
  \def\lo{0.12}    %
  \def\lext{0.1}  %
  \def\rext{1.05} %
  \def\yearLabel(#1,#2,#3){\node[above,black!60!cyan] at ({(#1-\yearOne)*\w/\n},{\lt*#2}) {#3};}

    \def\yearArrowLabel(#1,#2,#3,#4,#5){
    \def\xy{{(#1-\yearOne)*\w/\n}}; \pgfmathparse{int(#2*100)};
    \ifnum \pgfmathresult<0 %
      \def\yyp{{(\lt*(0.90+#2))}}; \def\yyw{{(\yyp-\lt*#3)}}
      \fill[color=#5,radius=2pt] (\xy,0) circle;
      \draw[<-,thick,color=#5,align=center]
        (\xy,\yyp) -- (\xy,\yyw)
        node[below,color=#5] at (\xy,\yyw) {\strut #4};
    \else %
      \def\yyp{{(\lt*(0.10+#2)}}; \def\yyw{{(\yyp+\lt*#3)}}
      \fill[color=#5,radius=2pt] (\xy,0) circle;
      \draw[<-,thick,color=#5,align=center]
        (\xy,\yyp) -- (\xy,\yyw)
        node[above] at (\xy,\yyw) {#4};
    \fi}
    
    \draw[->,thick] (-\w*\lext,0) -- (\w*\rext,0);
    
    \foreach \tick in {0,1,...,\n}{
      \def\x{{\tick*\w/\n}}
      \def\year{\the\numexpr \yearOne+\tick*1 \relax}
      \fill[black,radius=2.5pt] (\x,0) circle;
      \draw[thick] (\x,-0.0001) -- (\x,0.0001) %
	               node[below] {\year};
      \ifnum \tick<\n
        \foreach \ticko in {1,2,3,4,5,6,7,8,9,10,11}{
          \def\xo{{(\x+\ticko*\w/\n/12)}}
  	      \draw[thick] (\xo,0) -- (\xo,\lo);  %
	  }\fi
    }
    \draw[thick] (-1*\w/\n/12,0) -- (-1*\w/\n/12,\lo);
    \draw[thick] (-2*\w/\n/12,0) -- (-2*\w/\n/12,\lo);
    \draw[thick] ({\w+\w/\n/12},0) -- ({\w+\w/\n/12},\lo);
  
    \yearArrowLabel(2018.70,-1.5,0.5, $S^2$-GAN~\cite{wang2016generative}, cyan)
    \yearArrowLabel(2018.72, 0.5, 3.5, PrGAN~\cite{gadelha20173d}, cyan)  %
    \yearArrowLabel(2018.95,-1.5,1.5, DeepVoxel~\cite{sitzmann2019deepvoxels}, black!25!lime)  %
    \yearArrowLabel(2018.95,0.5,1.5,VON~\cite{zhu2018visual}, cyan)          
    \yearArrowLabel(2019.30,-1.5,3.5, HoloGAN~\cite{nguyen2019hologan}, violet)  %
    \yearArrowLabel(2019.50,0.5,1.5, SRN~\cite{sitzmann2019scene}\\\textcolor{orange}{GANSteerability~\cite{jahanian2020steerability}}, black!25!lime) 
    \yearArrowLabel(2019.75,-1.5,1.5, RGBD-GAN~\cite{noguchi2020rgbd}, red) 
    \yearArrowLabel(2019.95,0.5,2.5, DVR~\cite{niemeyer2020differentiable}, black!25!lime) 
    \yearArrowLabel(2020.05,-1.5,3.5, BlockGAN~\cite{nguyen2020blockgan}, violet) 
    \yearArrowLabel(2020.15,0.5,4.5, GANLatentDiscovery~\cite{voynov2020latent}, orange)
    \yearArrowLabel(2020.23,-1.5,1.5,NeRF~\cite{mildenhall2020nerf}, black!25!lime)
    \yearArrowLabel(2020.30,0.5,1.5, StyleRig~\cite{tewari2020stylerig}, red) 
    \yearArrowLabel(2020.40,-1.5,5.5, CONFIG~\cite{kowalski2020config}\\\textcolor{orange}{GANSpace~\cite{eric2020GANSpace})}\\\textcolor{orange}{InterFaceGAN~\cite{shen2020interpreting})}, red) 
    \yearArrowLabel(2020.55, 0.5,2.5, NGP~\cite{chen2021towards}\\\textcolor{orange}{SeFa~\cite{shen2021closedform}}, cyan)   %
    \yearArrowLabel(2020.60,-1.5,0.5, GRAF~\cite{schwarz2020graf}, violet)  %
    \yearArrowLabel(2020.65,-1.5,3.5, NeRF-W~\cite{martin2021nerf}, black!25!lime) 
    \yearArrowLabel(2020.75,0.5,1.5, PIE~\cite{tewari2020pie}, red) 
    \yearArrowLabel(2020.80,-1.5,4.5, NeRF++~\cite{zhang2020nerf++}, black!25!lime) 
    \yearArrowLabel(2020.90,-1.5,1.5, GIRAFFE~\cite{niemeyer2021giraffe}, violet)  %
    \yearArrowLabel(2020.93,0.5,2.5, pi-GAN~\cite{chan2021pigan}, violet)  %
    \yearArrowLabel(2020.93,4.5,0.5, PixelNeRF~\cite{yu2021pixelnerf}, black!25!lime) 
    \yearArrowLabel(2021.05,-1.5,7.0, GAN-Control~\cite{shoshan2021gan}, red)
    \yearArrowLabel(2021.15,0.5,1.5, NeRF--~\cite{wang2021nerf}, black!25!lime) 
    \yearArrowLabel(2021.30,-1.5,2.5, KiloNeRF~\cite{reiser2021kilonerf}\\Mip-NeRF~\cite{barron2021mipnerf}\\FastNeRF~\cite{garbin2021fastnerf}\\\textcolor{violet}{CAMPARI~\cite{niemeyer2021campari}}, black!25!lime) 
    \yearArrowLabel(2021.35,0.5,3.5, BARF~\cite{lin2021barf}, black!25!lime) 
    \yearArrowLabel(2021.35,5.5,0.5, VariTex~\cite{buhler2021varitex}, red) 
    \yearArrowLabel(2021.55,-1.5,0.5, Liao~\etal~\cite{liao2020towards}, violet)
    \yearArrowLabel(2021.65,0.5,0.5, ShadeGAN~\cite{pan2021shadegan}\\CIPS-3D~\cite{zhou2021cips3d}\\StyleNeRF~\cite{gu2022stylenerf}, violet)
    \yearArrowLabel(2021.92, 0.5,3.5, GOF~\cite{xu2021generative}\\\textcolor{black!25!lime}{URF~\cite{rematas2022urbannerf}}, violet) 
    \yearArrowLabel(2021.95, -6.2, 0.5, EG3D~\cite{chan2022efficient}\\GRAM~\cite{deng2022gram}\\StyleSDF~\cite{orel2022stylesdf}\\VolumeGAN~\cite{xu2022volumegan}, violet) 
    \yearArrowLabel(2022.10, -1.5, 0.5, \textcolor{cyan}{DepthGAN~\cite{shi20223d}}\\\textcolor{black!25!lime}{Block-NeRF~\cite{tancik2022block}}\\\textcolor{black!25!lime}{Instant-NGP~\cite{muller2022instant}}\\\textcolor{red}{DiscoFaceGAN~\cite{deng2020disentangled}}, violet) 
    \yearArrowLabel(2022.35, 0.5, 0.5, MVCGAN~\cite{zhang2022multi}\\Sem2NeRF~\cite{chen2022sem2nerf}\\Disentangled3D~\cite{tewari2022disentangled3d}, violet) 
    \yearArrowLabel(2022.45, -1.5, 5.5, VoxGRAF~\cite{schwarz2022voxgraf}\\EpiGRAF~\cite{skorokhodov2022epigraf}\\IDE-3D~\cite{sun2022ide3d}, violet) 
    \yearArrowLabel(2022.75, -1.5, 0.5, GMPI~\cite{zhao2022generative}\\SURF-GAN~\cite{kwak2022injecting}\\\textcolor{red}{3D-FM GAN~\cite{liu20223d}}, violet)  
    \yearArrowLabel(2022.90, 0.5, 3.5, 3D-SGAN~\cite{zhang20213d}\\RenderDiffusion~\cite{anciukevivcius2023renderdiffusion}, violet)  
    \yearArrowLabel(2022.95, -5.5, 0.5, 3DHumanGAN~\cite{yang20223dhumangan}\\DiffRF~\cite{muller2023diffrf}\\DiscoScene~\cite{xu2023discoscene}, violet)    
\end{tikzpicture}
}
\caption{\revise{Chronological overview of representative deep generative 3D-aware image synthesis methods which are categorized by different learning approaches. 
Methods in \textcolor{black!25!lime}{lime}, \textcolor{orange}{orange}, \textcolor{red}{red}, \textcolor{cyan}{cyan}, and \textcolor{violet}{violet}, are from~\cref{subsec:3d_novel_view_synthesis},~\cref{subsec:exploring_3d_control_in_2d_latent_space},~\cref{subsec:3d_parameter_as_controls},~\cref{subsec:3d_prior_as_constraints}, and~\cref{sec:3d_generative_models}, respectively. 
$S^2$-GAN~\cite{wang2016generative} and PrGAN~\cite{gadelha20173d} are published in 2016 and are not shown in scale. Best viewed in color. 
The INR-based novel view synthesis methods (refer to Section~\ref{subsec:3d_novel_view_synthesis}) are not the focus of this survey. They are included to offer background information relevant to the topics discussed. The papers represented in this figure are current up to December 2022.}
}
\label{fig:timeline} 
\end{figure*}
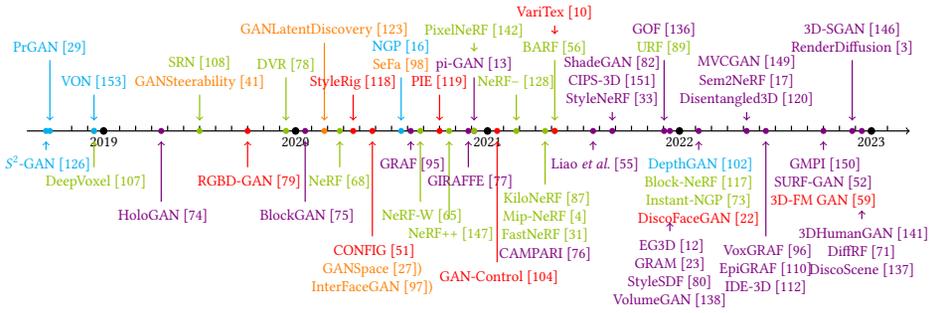